\documentclass[lettersize,journal]{IEEEtran}
\usepackage{amsmath,amsfonts}
\usepackage{algorithmic}
\usepackage{multirow}
\usepackage{tabularx}
\usepackage{booktabs}
\usepackage{algorithm}
\usepackage{makecell}
\usepackage{array}
\usepackage[caption=false,font=footnotesize,labelfont=rm,textfont=rm]{subfig}
\usepackage{textcomp}
\usepackage{stfloats}
\usepackage{url}
\usepackage{verbatim}
\usepackage{graphicx}
\usepackage{cite}
\usepackage{orcidlink}
\usepackage{bm}
\begin{document}

\title{CBMC-V3: A CNS-inspired Control Framework\\ Towards Agile  Manipulation with SNN}

\author{Yanbo Pang
	, Qingkai Li
	, Mingguo Zhao
\thanks{This research was supported by STI 2030-Major Projects 2021ZD0201402.}
}



\maketitle

\begin{abstract}
As robotic arm applications expand beyond traditional industrial settings into service-oriented domains such as catering, household, and retail, existing control algorithms struggle to achieve the level of agile manipulation required in unstructured environments characterized by dynamic trajectories, unpredictable interactions, and diverse objects. This paper presents a biomimetic control framework based on spiking neural network (SNN), inspired by the human central nervous system (CNS), to address these challenges. The proposed framework comprises five control modules—cerebral cortex, cerebellum, thalamus, brainstem, and spinal cord—organized into three hierarchical control levels (first-order, second-order, and third-order) and two information pathways (ascending and descending). All modules are fully implemented using SNN. The framework is validated through both simulation and experiments on a commercial robotic arm platform across a range of control tasks. The results demonstrate that the proposed method outperforms the baseline in terms of agile motion control capability, offering a practical and effective solution for achieving agile manipulation.
\end{abstract}


\section{Introduction}\label{sec:intro}
In recent years, robotic arms have been increasingly deployed in manipulation-intensive domains such as catering, household, and retail services, posing new challenges for control strategies. While a 7-DOF robotic arm can readily cut precise circular contours or transport heavy replacement parts, it still struggles with delicate tasks such as peeling an apple—an operation that the human arm performs effortlessly. This gap largely arises from humans’ strong motion agility, which enables robust and flexible handling of delicate, contact-rich interactions. However, defining and quantifying agility in motion control remains challenging due to its context-dependent and multifaceted nature.

Drawing insights from \cite{sheppard2006agility,8610238}, we characterize agile motion control using four key features: (1) precision, (2) adaptation, (3) resilience, and (4) energy efficiency. Precision measures an algorithm’s ability to accurately track a desired trajectory, for example in terms of end-effector position and orientation tracking errors. Adaptation evaluates the ability to maintain tracking precision across different tasks, such as under varying load masses and trajectory configurations. Resilience reflects how quickly the algorithm recovers from unstructured interactions or disturbances. Energy efficiency quantifies the computational resources required by the algorithm.

Control algorithms that combine Whole-Body Control (WBC) \cite{Abe2007, liu2011interactive, weighttans2011, Lee2012Intermediate, Escande2014, Kim2019, han2021recursive} and Model Predictive Control (MPC) \cite{carron2019data, pankert2020perceptive, wang2023trajectory} can achieve high precision in well-defined environments and typically exhibit high energy efficiency. However, their resilience is often compromised as model complexity increases. Moreover, these methods lack effective mechanisms to cope with environmental variations and model inaccuracies, such as objects with differing shapes or masses and elastic or biomimetic joints, resulting in poor adaptation.

Artificial neural network (ANN) can learn and adapt to environmental changes, resulting in relatively strong adaptation capabilities. When combined with model-based dynamics, it can also achieve relatively high precision. For example, \cite{chaoui2009ann} designed feedforward and feedback neural networks to compensate for joint flexibility and friction; \cite{he2017adaptive} employed fuzzy neural networks with adaptive impedance control to handle unknown dynamics and state constraints; \cite{wang2018adaptive} proposed a neural-network-based adaptive controller capable of addressing nonlinear dynamics and external disturbances without prior models; \cite{salloom2020adaptive} integrated genetic algorithm observers with neural networks to compensate for underwater disturbances; \cite{liu2019adaptive} investigated the role of hidden-layer size in adaptive control and proposed dynamic adjustment strategies; and \cite{pham2020adaptive} applied radial basis function networks with dynamic surface control for robust dual-arm coordination. However, these methods typically exhibit limited resilience and energy efficiency due to the computational characteristics of ANN \cite{bing2018survey}. Moreover, the limited ability of ANN to model temporal dynamics further restricts their adaptation in highly dynamic manipulation tasks.

Recently, spiking neural network (SNN) combined with cerebellum-inspired architecture has emerged as promising alternatives. By emulating biological neurons, SNN processes and stores information through spike trains and membrane potential dynamics, offering significant advantages in energy efficiency \cite{bing2018survey}. \cite{carrillo2008real} developed a cerebellum-inspired SNN trained via spike-timing-dependent plasticity (STDP) to learn robotic arm dynamics for torque control; \cite{abadia2019robot} extended this model to human–robot collaboration scenarios, demonstrating robustness under unknown disturbances; \cite{abadia2021cerebellar} addressed control delays using this network, highlighting its potential for teleoperation and cloud robotics; and \cite{doi:10.1126/scirobotics.adp2356} further refined the framework to enable robotic arms to dynamically adjust compliance. By modeling human neural processing at the level of individual neurons and closely mimicking the structure of the human cerebellum, these approaches can exhibit human-like behavior and strong adaptation capabilities. However, despite the low per-neuron energy consumption, the extremely large network scale (e.g., over 60k neurons and 32M synaptic connections) limits the overall energy efficiency of the framework. Moreover, the use of unsupervised, biologically inspired learning rules such as STDP—whose convergence properties lack theoretical guarantees—constrains achievable precision and resilience.

In our previous work CBMC-V2 \cite{10769842}, we presented a hierarchical control framework inspired by the central nervous system (CNS). Compared with cerebellum-inspired SNN approaches, this framework achieves higher energy efficiency by modular design and neuromorphic hardware deployment, and attains improved precision and resilience by incorporating gradient-based learning into both the cerebral cortex and cerebellum modules. However, the framework exhibits weak adaptation due to limited learning capability, arising from insufficient sensory inputs (e.g., load information), the absence of SNN in some modules, and the tight coupling between control and planning in the cerebral cortex module. Furthermore, because the cerebellum module largely preserves the network structure adopted in \cite{abadia2019robot, abadia2021cerebellar, doi:10.1126/scirobotics.adp2356}, the framework inherits their inherent limitations, which constrain the achievable precision, resilience, and energy efficiency.

Overall, this research aims to provide a motion control framework towards agile manipulation by addressing the limitations identified in CBMC-V2. To this end, we propose CBMC-V3, a CNS-inspired, fully SNN-based control framework. The main contributions of this work are threefold: (1) the proposal of a CNS-inspired control framework; (2) the design of a fully SNN-based implementation; and (3) the demonstration of agile motion control in both simulation and on a commercial robotic arm.

The remainder of this paper is organized as follows: Section \ref{sec:framework} introduces the framework structure; Section \ref{sec:module} details the SNN-based implementations of individual modules; Section \ref{sec:simulation} presents simulation results; Section \ref{sec:experiment} reports experiment results on a commercial robotic arm, and Section \ref{sec:conclusion} concludes this paper.
\begin{figure*}[!t]
	\centering
	\subfloat[]{\includegraphics[width=0.3\linewidth]{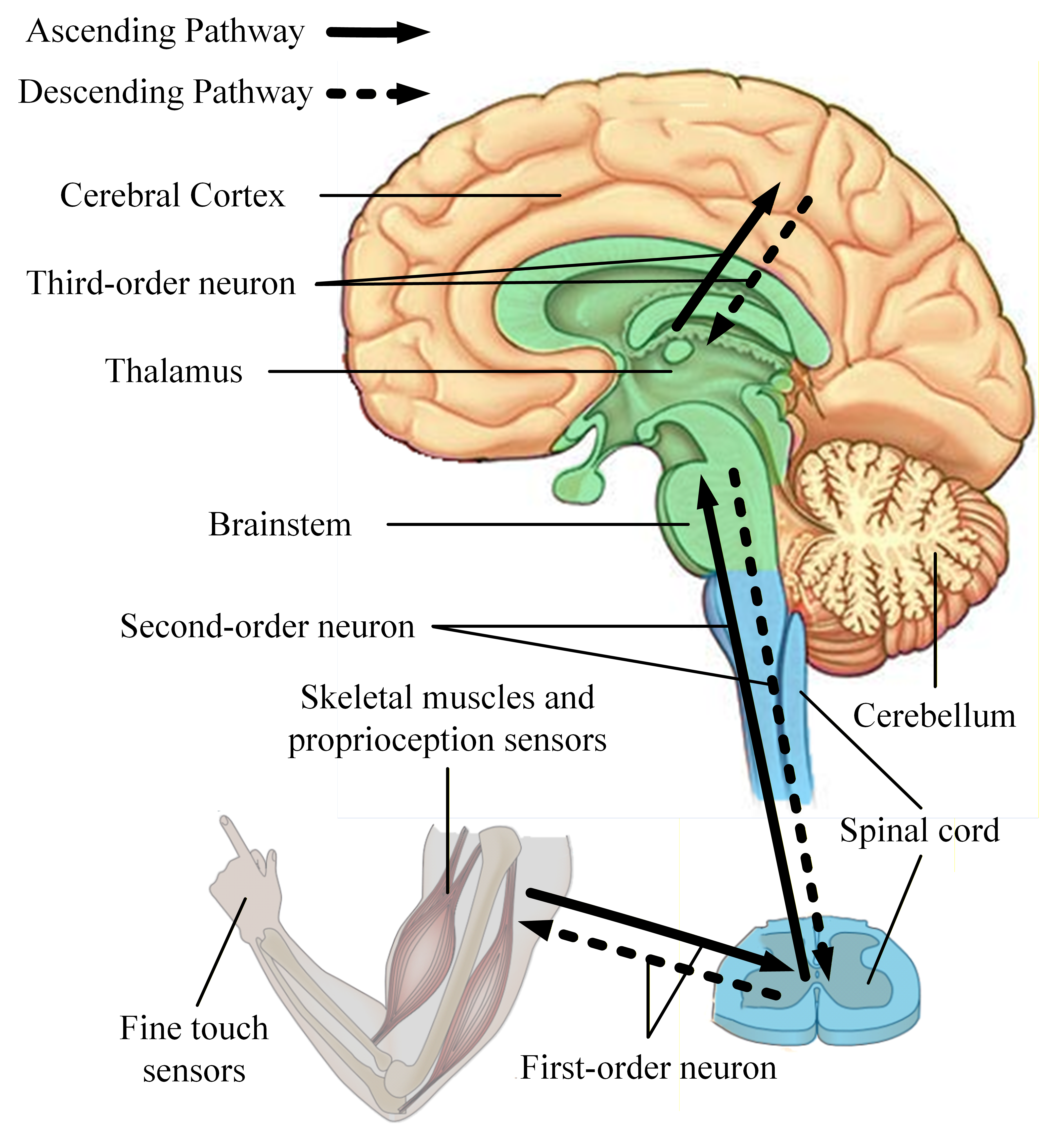}%
		\label{fig:framework1}}
	\hfil
	\subfloat[]{\includegraphics[width=0.295\linewidth]{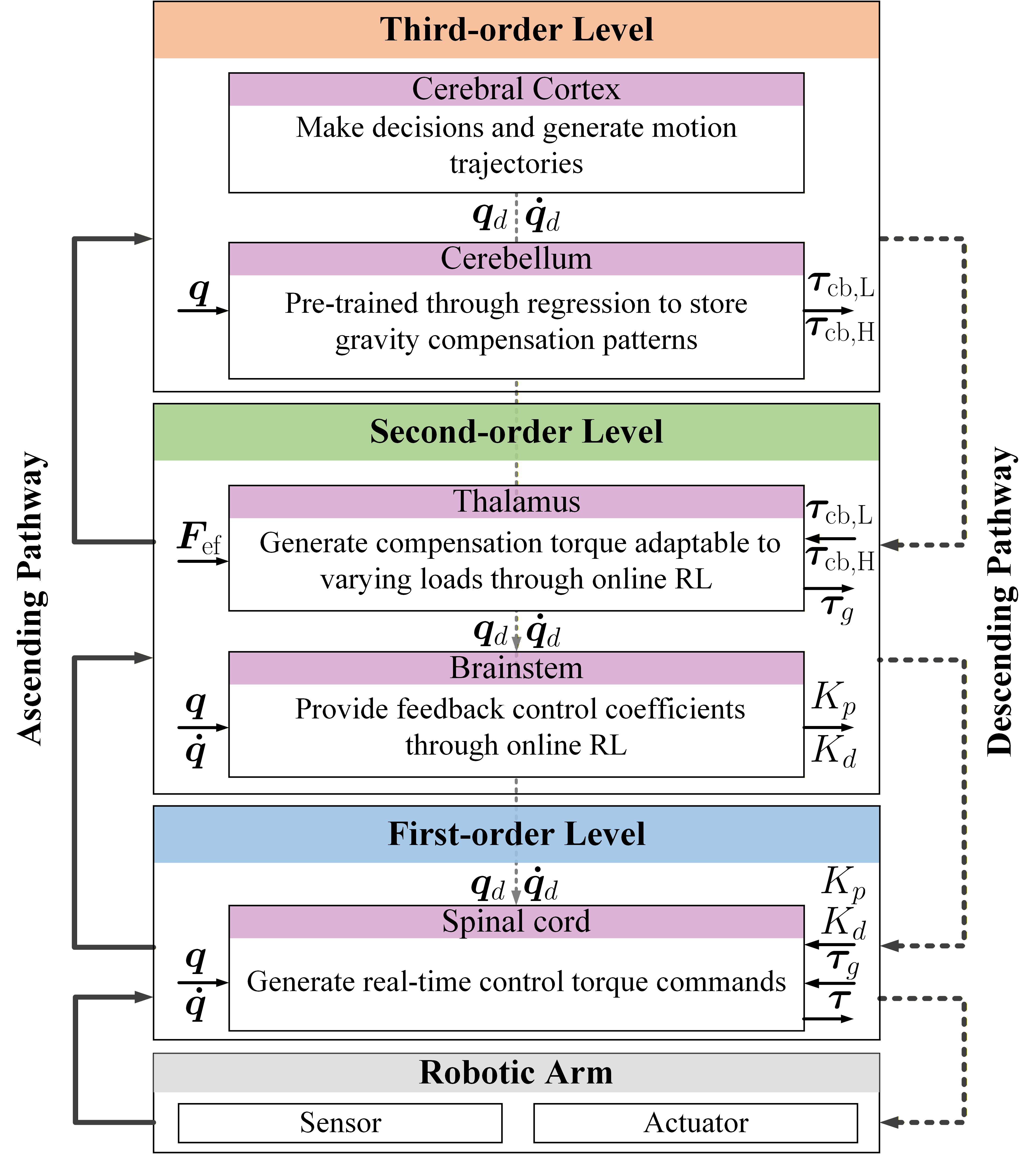}%
		\label{fig:framework2}}
	\hfil
	\subfloat[]{\includegraphics[width=0.36\linewidth]{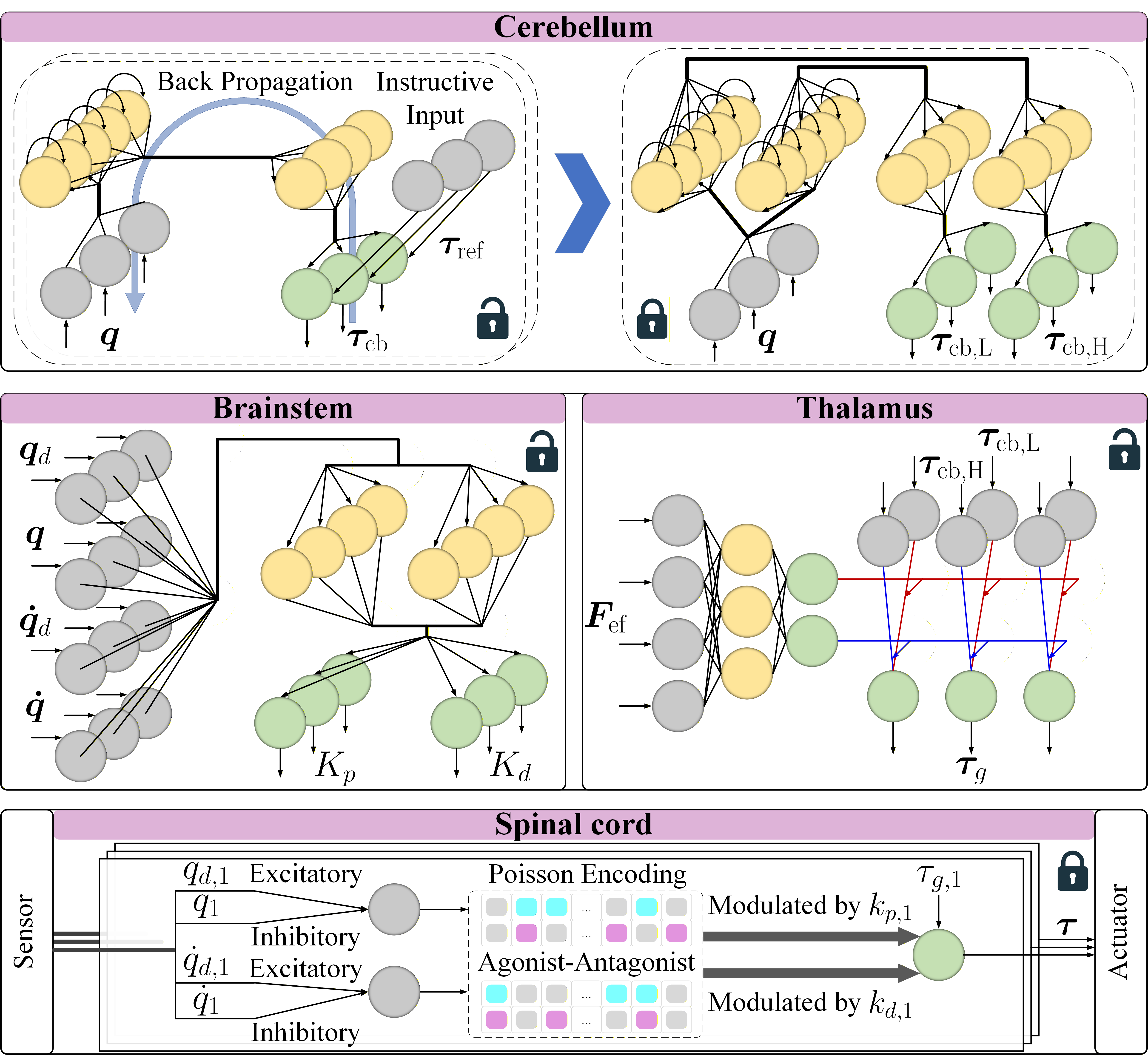}%
		\label{fig:framework3}}
	\caption{Overview of the human CNS and the proposed framework. (a) Human CNS relating to motor control. (b) Diagram of our proposed framework, featuring five modules, three control levels and two pathways. (c) Neural network design of each module with three types of spiking neurons: connecting fiber (grey), LIF neuron (yellow) and non-spiking LIF neuron (green).}
	\label{fig:framework}
\end{figure*}

\section{Framework structure}\label{sec:framework}

Inspired by the motor control mechanism of the human CNS, as illustrated in Fig.\ref{fig:framework1}, the structure of our proposed framework CBMC-V3 is shown in Fig.\ref{fig:framework2}. This framework consists of three key aspects: (1) five modules including the cerebral cortex module, the cerebellum module, the thalamus module, the brainstem module, and the spinal cord module; (2) three hierarchical control levels, namely the first-order level, second-order level, and third-order level; (3) two information pathways, i.e. the ascending pathway and the descending pathway. In the following sub-sections, we will introduce each of these aspects and present the pseudocode implementation of this framework.

\subsection{Modules}

\subsubsection{The cerebral cortex module}
The cerebral cortex module makes decisions and generates motion trajectories. In this research, it stores and replays pre-planned velocity $\bm{\dot{q}}_d\in\mathbb{R}^n$ and position $\bm{q}_d\in\mathbb{R}^n$ of the $n$ actuated joints.

\subsubsection{The cerebellum module}
The cerebellum module is a cerebellar-like network pre-trained to store gravity compensation motor patterns for corresponding load masses. In human motor control, the cerebellum is responsible for coordinating and regulating body movements, ensuring muscle and postural balance. It integrates sensory inputs, such as muscle tension and joint position, and adjusts the magnitude and direction of motor commands to achieve precise and fluid motion \cite{LLOBERA2023190}. The gravity compensation torque for a robotic arm is designed to counteract the gravitational forces acting on the arm itself. It plays a crucial role in ensuring the smoothness and stability of motion. The cerebellum can be regarded as performing a function similar to gravity compensation, and the cerebellum module in our framework is therefore designed to provide this compensation.

Furthermore, we notice that during human manipulation, accurate dynamic model is not required. Instead, the brain relies on stored motor patterns to cope with different load conditions. For instance, when grasping an object, we estimate its mass based on prior experience and adjust the exerted force accordingly. Such knowledge is typically not acquired on site during execution, but rather accumulated over time through long-term daily interactions. Inspired by this biological mechanism, we equip the cerebellum module with the ability to pre-learn $m$ gravity compensation patterns corresponding to different load masses. These patterns can then be retrieved and combined in prediction to generate a compensation torque matrix
\begin{equation}\label{eq:cb_cbmc}
	\begin{split}
		\bm{T}_{{\text{cb}}}(\bm{q}) &= (\bm{\tau}_{\text{cb},1}(\bm{q}),\bm{\tau}_{\text{cb},2}(\bm{q}),\cdots,\bm{\tau}_{\text{cb},m}(\bm{q}))\\
		&=\begin{pmatrix}
			\tau_{\text{cb},11} & \cdots & \tau_{\text{cb},1m}\\
			\vdots & \ddots & \vdots \\
			\tau_{\text{cb},n1} & \cdots & \tau_{\text{cb},nm}
		\end{pmatrix}
	\end{split}
\end{equation}
as a function of actual joint positions $\bm{q}\in\mathbb{R}^n$, where $\tau_{\text{cb},ji}(\bm{q}), j=1,\cdots,n, i=1,\cdots,m$ denotes the torque provided for the $j$-th joint under the $i$-th compensation pattern.

\subsubsection{The thalamus module}
The thalamus module learns to generate compensation torque adaptable to arbitrary loads by online RL using proprioceptive information and the  cerebellum module output. In unstructured environments, the selection of motor patterns may be inaccurate due to factors such as visual deception. For instance, the mass of an opaque box cannot be reliably estimated visually, which may lead to insufficient force to move the load or, conversely, an excessive initial force causing large errors. Nevertheless, humans can quickly adjust motor output upon contacting the object, using sensory feedback to correct their movement patterns. This phenomenon is closely associated with the function of the thalamus, which serves as a central hub for motor coordination and plays a critical role in multimodal perception and sensorimotor integration \cite{LLOBERA2023190,Halassa_2022}. 

Analogously, a robotic arm can perceive the load mass via force sensors installed at the end-effector. The thalamus module, similar to humans, uses the sensed load information to appropriately integrate the torque matrix provided by the cerebellum module, i.e., 
\begin{equation}\label{eq:tm_cbmc}
	\begin{split}
		\bm{\tau}_g(\bm{q},\bm{F}_\text{ef}) &= \bm{T}_\text{cb}(\bm{q})\cdot \bm{w}(\bm{F}_\text{ef})\\
		&= \begin{pmatrix}
			\tau_{g,1}\\
			\vdots \\
			\tau_{g,n}
		\end{pmatrix}
		=\begin{pmatrix}
			w_1\tau_{\text{cb},11} +\cdots+ w_m\tau_{\text{cb},1m}\\
			\vdots \\
			w_1\tau_{\text{cb},n1} +\cdots+ w_m\tau_{\text{cb},nm}
		\end{pmatrix}
	\end{split}
\end{equation}
where $\bm{F}_\text{ef}$ denotes the end-effector torque sensor feedback, $\bm{w}(\bm{F}_\text{ef})=(w_1,w_2,\dots,w_m)^\top$ represents the $m$ dynamically learned weight coefficients of the thalamus module, and $\bm{\tau}_g(\bm{q},\bm{F}_\text{ef}) = (\tau_{g,1},\dots,\tau_{g,n})^\top$ are the resulting gravity compensation torque of each joint for arbitrary load mass. 

By combining the prior knowledge stored in the cerebellum module with the adaptive weighting capability of the thalamus module, the framework effectively addresses the unknown dynamics of the robotic arm’s end-effector load. 

\subsubsection{The brainstem module}\label{sec:bs}
The brainstem module learns to provide appropriate feedback control coefficients by online RL using proprioceptive information. The brainstem has been shown to contain pathways involved in movement termination and reorientation, and to interact with the spinal cord to coordinate motor behaviors \cite{merel2019hierarchical,annurev:/content/journals/10.1146/annurev-neuro-082321-025137}. Motivated by this, the brainstem module in our framework dynamically adjusts the proportional and derivative coefficients of the $n$ actuated joints based on trajectory information, i.e., 
\begin{align}\label{eq:bs_cbmc}
	diag(k_{p,1},k_{p,2},\cdots,k_{p,n}) = \bm{K}_P(\bm{\dot{q}}_d,\bm{q}_d,\bm{\dot{q}},\bm{q})\\
	diag(k_{v,1},k_{v,2},\cdots,k_{v,n}) = \bm{K}_V(\bm{\dot{q}}_d,\bm{q}_d,\bm{\dot{q}},\bm{q})
\end{align}
where $\bm{\dot{q}}\in\mathbb{R}^n$ is the actual velocity of the $n$ actuated joints.

\subsubsection{The spinal cord module}\label{sec:sc}
The spinal cord module combines feedback control coefficients from the brainstem module and the compensation torque from the thalamus module to generate real-time control torque commands. 

In the human CNS, the spinal cord can generate reflexive actions independently via reflex arcs, enabling rapid responses to sensory stimuli without cortical input \cite{LLOBERA2023190}. Inspired by this mechanism, the spinal cord module produces torque commands at real-time control frequency with a PD-like feedback control mechanism, i.e., 
\begin{equation}\label{eq:pd_cbmc}
	\begin{aligned}
		\bm{\tau}_{\text{fb}}(\bm{\dot{q}}_d,\bm{q}_d,\bm{\dot{q}},\bm{q})&=\bm{K}_V(\bm{\dot{q}}_d,\bm{q}_d,\bm{\dot{q}},\bm{q})(\bm{\dot{q}_d}-\bm{\dot{q}})\\
		&+
		\bm{K}_P(\bm{\dot{q}}_d,\bm{q}_d,\bm{\dot{q}},\bm{q})(\bm{q_d}-\bm{q})
	\end{aligned}
\end{equation}
By integrating spinal feedback control with brainstem-driven adaptive coefficients modulation, the framework effectively addresses non-linearity, Coriolis and centrifugal forces from unknown trajectories.

This reflexive torque is then summed with the compensation torque from the thalamus to generate control torque command $\bm{\tau}$ for actuators, i.e., 
\begin{equation}\label{eq:cbmc}
	\bm{\tau}=\bm{\tau}_{\text{fb}}(\bm{\dot{q}}_d,\bm{q}_d,\bm{\dot{q}},\bm{q})+\bm{\tau}_g(\bm{q},\bm{F}_\text{ef})=(\tau_1,\cdots,\tau_n)^\top
\end{equation}


\subsection{Control levels}
Different neural regions in the CNS operate hierarchically during motor control, with each level specializing in particular tasks while communicating and cooperating through neural pathways. This hierarchical organization enables organisms to execute a wide range of motion tasks flexibly and stably in unstructured environments.

Physiological studies indicate that the multilevel transmission of neural signals is a key component of this hierarchy. Proprioceptive information is conveyed from sensory receptors to the brain via a three-order neuronal pathway\cite{LLOBERA2023190, moore2018clinically, sinnatamby2013last}, as illustrated in Fig.\ref{fig:framework1}. First-order neurons receive impulses from the skin and muscles and transmit them to the spinal cord. Some of this information is processed locally in the spinal cord and then relayed back to the muscles through these first-order neurons. The remaining signals pass through second-order neurons, which transmit impulses via the brainstem to the thalamus. Finally, third-order neurons carry these impulses to the cerebral cortex and cerebellum, and are responsible for information exchange between them.

Inspired by this three-order neuronal pathway, the proposed framework implements a corresponding three-level control hierarchy operating at different frequencies:
\begin{itemize}
	\item First-order level: Comprising the spinal cord module, this level operates at the highest frequency. It generates real-time control torques sent to the robotic arm, ensuring accurate and rapid responses.
	
	\item Second-order level: Including the brainstem module and thalamus module, this level operates at an intermediate frequency. It serves as a relay between the first-order and third-order levels, performing two main functions: regulating the feedback control coefficients of the first-order level, and integrating and filtering the torques provided by the third-order level to supply gravity compensation torque to the first-order level.
	
	\item Third-order level: Comprising the cerebellum module and cerebral cortex module, this layer operates at the lowest frequency. It generates motion trajectories and provide  compensation torques based on motor patterns, maintaining movement stability and smoothness.
\end{itemize}

\subsection{Information pathways}
As illustrated in Fig.\ref{fig:framework1}, the human CNS employs ascending and descending pathways for sensory information transmission and motor command execution, respectively\cite{sinnatamby2013last, moore2018clinically}. The ascending pathway conveys peripheral sensory information to the brain for perception and integration, while the descending pathway transmits motor commands from the brain to the spinal cord and peripheral nerves to control movements. Following this principle, our framework incorporates two analogous information pathways:
\begin{itemize}
	\item Ascending pathway: This pathway transmits proprioceptive and touch information, namely joint positions, velocities, and end-effector torque feedback. It simulates the process by which muscle receptors relay signals to the CNS for processing and can be regarded as the input process of the framework.
	
	\item Descending pathway: This pathway transmits joint torque commands and module coefficients. It simulates the process by which the CNS delivers motor commands to muscle effectors and can be regarded as the output process of the framework.
\end{itemize}

\subsection{Pseudocode implementation}
To clarify the interplay among modules, levels, and pathways illustrated in Fig.~\ref{fig:framework2}, the pseudocode implementation of this framework is provided in Algorithm \ref{alg:cbmc}. 

Information travels different length to reach different modules along the pathway, so the framework is organized into three control levels with different frequencies.

On the ascending side, proprioceptive data (actual joint states $(\bm{\dot{q}},\bm{q})$ and desired joint states $(\bm{\dot{q}}_d,\bm{q}_d)$) are transmitted via first-order neurons to the spinal cord module. Part of this input is locally processed to generate feedback torque $\bm{\tau}_{\text{fb}}$; part is relayed via second-order neurons to the brainstem module, which adaptively provides feedback coefficients $(\bm{K}_P,\bm{K}_V)$; and part is sent via third-order neurons to the cerebellum module, which outputs compensation matrix  $\bm{T}_{\text{cb}}$. 

On the descending side, $\bm{T}_{\text{cb}}$ is conveyed via third-order neurons to the thalamus module. It uses end-effector torque feedback $\bm{F}_\text{ef}$ to generate a weight vector $\bm{w}$ which filters $\bm{T}_{\text{cb}}$ to produce the gravity compensation torque $\bm{\tau}_g$. This torque is passed via second-order neurons to the spinal cord module. In the spinal cord module, $\bm{\tau}_g$ is combined with $\bm{\tau}_{\text{fb}}$ to form the control torque command $\bm{\tau}$, which is delivered through first-order neurons to the actuators.

\begin{algorithm}
	\caption{CBMC-V3}
	\label{alg:cbmc}
	\small
	\begin{algorithmic}[1]
		\REQUIRE actual joint velocity and position $\bm{\dot{q}},\bm{q}$, planned joint velocity and position $\bm{\dot{q}}_d,\bm{q}_d$, end-effector torque sensor data $\bm{F}_{ef}$
		\ENSURE joint control torque $\bm{\tau}$\\\hspace{1mm}
		
		\STATE Main thread:
		\STATE initialize first-order, second-order and third-order level control cycle $T_1,T_2,T_3$
		\STATE initialize trajectory variables $\bm{\dot{q}}_d,\bm{q}_d,\bm{\dot{q}},\bm{q}$
		\STATE initialize end-effector torque variable $\bm{F}_{ef}$
		\STATE initialize control torque variable $\bm{\tau}$
		\STATE initialize intermediate variables $\bm{K}_V,\bm{K}_P,\bm{\tau}_g,\bm{T}_{{\text{cb}}}$
		\STATE read trajectory length $L$
		
		\FOR {$t=1$ to $L$}
		
		\STATE read data from sensors and update $\bm{\dot{q}}_d,\bm{q}_d,\bm{\dot{q}},\bm{q},\bm{F}_{ef}$
		\IF{$t$ Mod $T_1==0$}
		\STATE run sub-thread 1
		\ENDIF
		\IF{$t$ Mod $T_2==0$}
		\STATE run sub-thread 2
		\STATE run sub-thread 3
		\ENDIF
		\IF{$t$ Mod $T_3==0$}
		\STATE run sub-thread 4
		\ENDIF
		\STATE send $\bm{\tau}$ to actuators
		
		\ENDFOR \\\hspace{1mm}
		
		\STATE Sub-thread 1:
		\STATE read $\bm{\dot{q}}_d,\bm{q}_d,\bm{\dot{q}},\bm{q},\bm{K}_V,\bm{K}_P,\bm{\tau}_g$
		\STATE process through the spinal cord module
		\STATE update $\bm{\tau}$\\\hspace{1mm}
		
		\STATE Sub-thread 2:
		\STATE read $\bm{\dot{q}}_d,\bm{q}_d,\bm{\dot{q}},\bm{q}$
		\STATE process through the brainstem module
		\STATE update $\bm{K}_V,\bm{K}_P$\\\hspace{1mm}
		
		\STATE Sub-thread 3:
		\STATE read $\bm{T}_{{\text{cb}}},\bm{F}_{ef}$
		\STATE process through the thalamus module
		\STATE update $\bm{\tau}_g$\\\hspace{1mm}
		
		\STATE Sub-thread 4:
		\STATE read $\bm{q}$
		\STATE process through the cerebellum module
		\STATE update $\bm{T}_{{\text{cb}}}$
		
	\end{algorithmic}
\end{algorithm}


\section{Module design}\label{sec:module}
Building on the framework introduced in the previous section, this chapter focuses on the implementation of each module using SNNs. The cerebral cortex module, which serves only for trajectory storage and playback, is therefore excluded from the discussion. All implementations in this chapter target a 7-DOF arm, i.e., $n=7$.

\subsection{Spiking neuron}
SNN employs spiking neurons as its computation units. A variety of spiking neuron models have been proposed in the literature, such as the Hodgkin–Huxley model \cite{hodgkin1952quantitative} and the Leaky Integrate-and-Fire (LIF) model along with its derivatives \cite{burkitt2006review}. The LIF model is of particular interest here, since it not only captures the dynamic process of membrane potential accumulation and leakage, but also maintains relatively high computational efficiency \cite{stein1965theoretical}.

The input current signal generated by a spike train $S(t)=\sum_f s(t-t_f),f=1,2,\cdots$ acting on a neuron can be described as 
\begin{equation}
	i\left(t\right)=\int_{0}^{\infty}{S\left(s-t\right)\exp\left(-s/\tau_\text{s}\right)\text{d}s}
\end{equation}
where $\tau_\text{s}$ denotes the synaptic time constant. The membrane potential $u(t)$ of the neuron evolves according to the following dynamics: 
\begin{equation}\label{lif}
	\tau_\text{mb}\frac{\text{d}u}{\text{d}t}=u_{\text{reset}}-u\left(t\right)+R\left(i_0\left(t\right)+\sum w_ji_j\left(t\right)\right)
\end{equation}
where $\tau_\text{mb} = RC$ is the membrane time constant determined by the membrane resistance $R$ and capacitance $C$. Here, $u_{\text{reset}}$ is the reset potential after firing, $i_0(t)$ represents the external current driving the neuron, $i_j(t)$ denotes the input current from the $j$-th synapse, and $w_j$ is the synaptic weight associated with the $j$-th synapse. When the membrane potential $u$ reaches a certain threshold $u_{\text{fire}}$, the neuron emits a spike and the potential is reset to $u_{\text{reset}}$. Therefore, the operational process of a spiking neuron can be interpreted as follows:

1. When the membrane potential $u(t)$ exceeds the firing threshold $u_{\text{fire}}$, the neuron emits a spike $s(t)$, expressed as
\begin{equation}
	s(t) = \Theta(u(t) - u_{\text{fire}})
\end{equation}
where
\begin{equation}\label{actifunc}
	\Theta(x) = \begin{cases}
		1, & x\geq 0 \\
		0, & x < 0
	\end{cases}
\end{equation}
is the Heaviside step function. Simultaneously, the membrane potential is reset to $u_{\text{reset}}$. When the membrane potential $u(t)$ remains below the threshold $u_{\text{fire}}$, no spike is generated.

2. Update equation (\ref{lif}).

3. Return to Step 1.


In our framework, we adopt a back-propagation-based training method\cite{8891809} for SNN. Since the activation function of the neuron in equation (\ref{actifunc}) is a discontinuous function, directly computing its derivative can lead to network instability. Consequently, various surrogate gradient methods have been proposed \cite{8891809}. A common approach is to use the original step function $\Theta(x)$ during the forward pass, while replacing its derivative with $\sigma'(x)$ instead of $\Theta'(x)$ during back propagation, where $\sigma(x)$ is referred to as the surrogate activation function. Typically, $\sigma(x)$ is a smooth and continuous function that resembles the shape of $\Theta(x)$ while enabling stable gradient computation.

As illustrated in Fig.\ref{fig:framework3}, the framework employs three types of neurons across all modules, represented in gray, yellow, and green. The gray neurons correspond to input fibers, which serve as the data interface for each module via connections to the ascending or descending pathway. The yellow neurons represent LIF neurons, which constitute the core units of SNNs and are responsible for information storage, processing, and learning. The green neurons correspond to non-spiking LIF neurons, implemented by setting the firing threshold $u_\text{fire}$ of standard LIF neurons to infinity. Unlike spiking neurons, their output is the continuous membrane potential $u(t)$ rather than discrete spikes, thereby enabling spike-to-continuous signal conversion. These neurons, connecting to the descending pathway, serve as the decoding layer of each module.

\subsection{Cerebellum module}\label{cbm}
Humans are capable of rapidly acquiring complex skills through imitation, and neural networks exhibit similar properties. Therefore, compared with unsupervised learning approaches, training a neural network to perform gravity compensation via imitation provides a faster and more efficient solution. Under the condition of a known robotic arm dynamic model, the gravity compensation torque represents a nonlinear mapping from joint positions to joint torques. In the machine learning domain, the task of fitting a curve to observed data under supervision is referred to as a regression task, for which extensive research and mature methods already exist\cite{henkes2024spiking,robotics13090126}. 

\begin{figure}[h]
	\centering
	\subfloat[]{\includegraphics[width=0.49\linewidth]{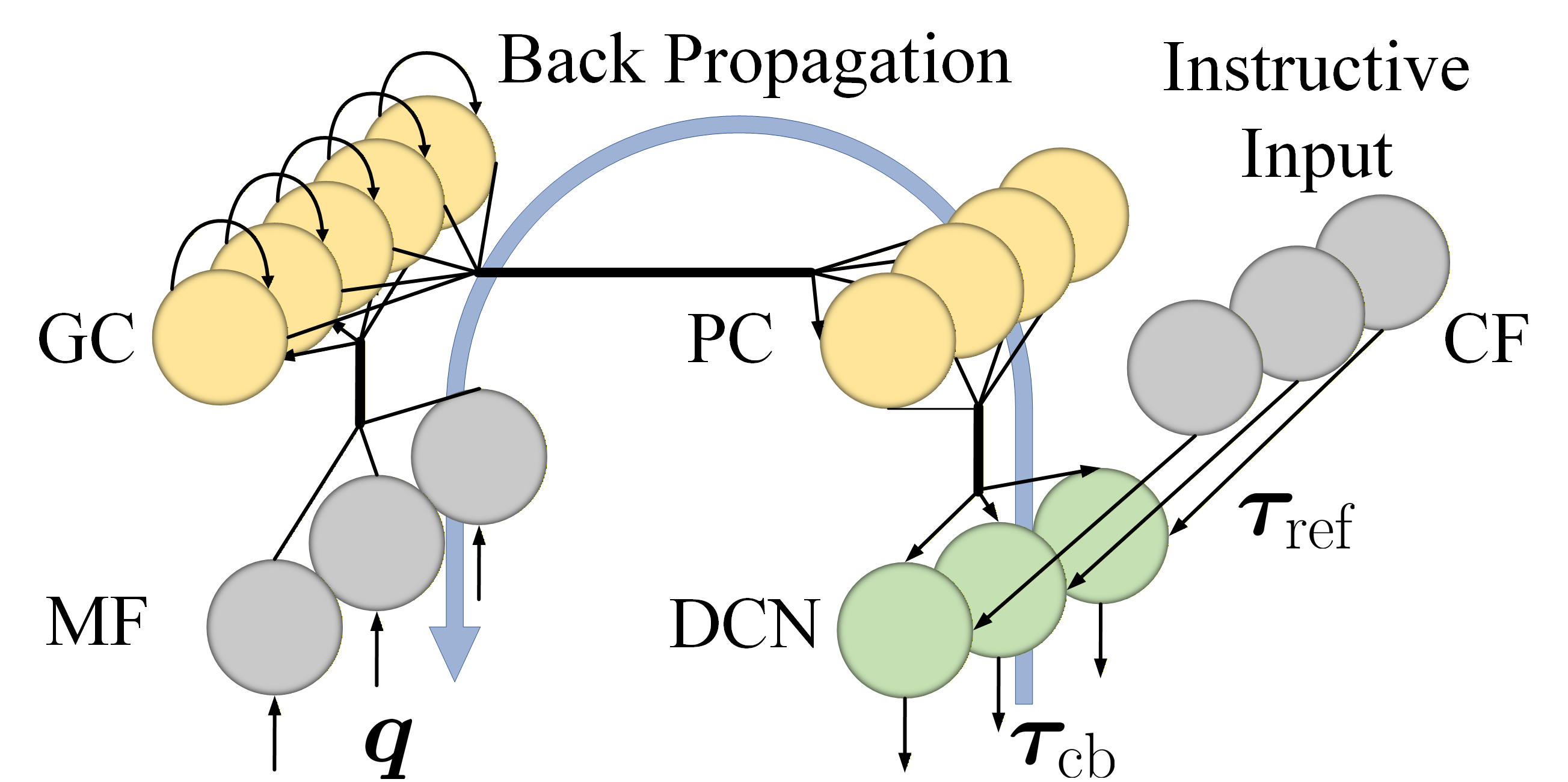}%
		\label{fig:cerebellum_train}}
	\hfil
	\subfloat[]{\includegraphics[width=0.49\linewidth]{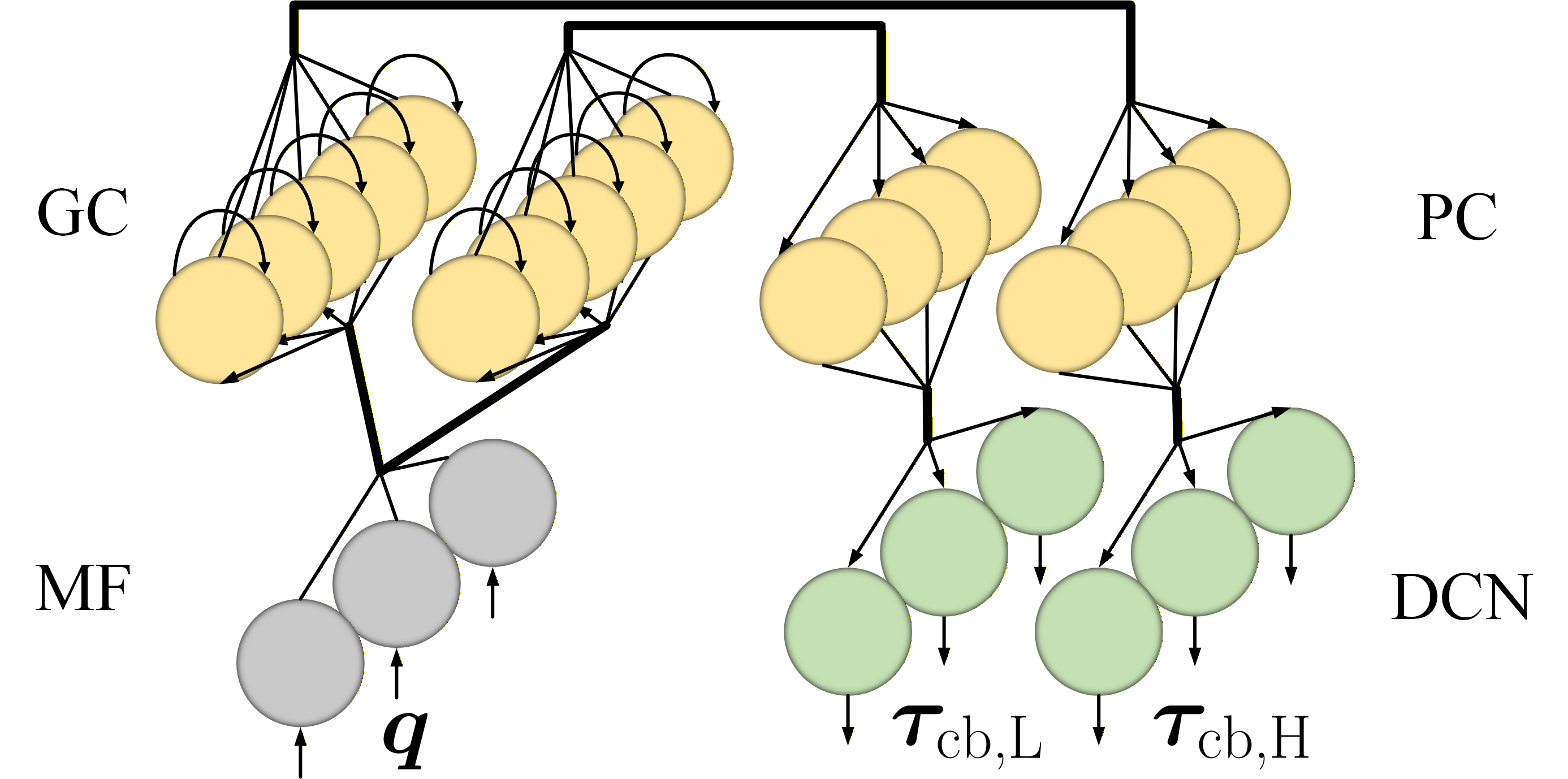}%
		\label{fig:cerebellum_predict}}
	\caption{Structure of the cerebellum module. (a) In training. (b) In prediction.}
	\label{fig:cerebellum}
\end{figure}

As shown in Fig.\ref{fig:cerebellum_train}, to implement the cerebellum module, a four-layer recurrent SNN inspired by \cite{abadia2019robot,abadia2021cerebellar} is designed to store one motor pattern corresponding to a particular load mass. The input variable $\bm{q} \in \mathbb{R}^7$ corresponds to the actual joint positions of the robotic arm, and the output $\bm{\tau}_\text{cb} \in \mathbb{R}^7$ represents the gravity compensation torque under a given load condition.

Variables from the ascending pathway first arrive at the mossy fiber (MF) layer, which is fully connected to the granule cell (GC) layer consisting of 300 LIF neurons. The GC layer, in turn, is fully connected to the Purkinje cell (PC) layer containing 200 LIF neurons. Each PC neuron receives input from both the current and previous time step of the GC layer, capturing temporal dynamics. The output of the PC layer is fully connected to the deep cerebellar nuclei (DCN) layer, which consists of 140 non-spiking LIF neurons. The decoding of the DCN layer is inspired by population coding in the animal motor cortex \cite{georgopoulos1986neuronal}. During primate arm movements, individual motor cortex neurons do not directly represent movement direction; rather, the movement direction emerges from a weighted vector sum over the preferred directions of a neuronal population. Analogously, the 140 DCN neurons are grouped in sets of 20, and their membrane potentials are averaged to obtain a 7-dimensional torque output $\bm{\tau}_\text{cb}$, i.e.,
\begin{equation}
	\tau_{\text{cb},i}=\sum_{j=20\cdot(i-1) + 1}^{20\cdot i}u_{\text{DCN},j}
\end{equation}
where $\tau_{\text{cb},i}, i=1,\cdots,7$ represents the torque for each joint and $u_{\text{DCN},j}, j=1,\cdots,140$ denotes the membrane potential of each DCN neuron. The network employs rate-based encoding\cite{gerstner2002spiking}, with a time window of length $T$.

The training data $\bm{\tau}_\text{ref}$ are generated using the robotic arm’s dynamic model. Under the condition that the training data adequately cover the workspace, this regression-based approach is sufficient to achieve effective gravity compensation. The network is trained in a supervised manner, with the objective of minimizing the error between the network output and the reference torque. Structurally, this is reflected in the DCN layer neurons receiving both excitatory inputs from the PC layer neurons and inhibitory inputs from the climbing fiber (CF) neurons. Mathematically, the network output torque $\bm{\tau}_\text{cb}$ is compared with the reference torque $\bm{\tau}_\text{ref}$ to obtain an error, which is quantified using the mean squared error (MSE Loss), i.e., 
\begin{equation}
	\text{MSE}=\frac{\sum_{i=1}^{7}(\tau_\text{cb,i}-\tau_\text{ref,i})^2}{7}
\end{equation}  
The weights are then updated via back propagation, thereby completing the training process. Detailed training procedure is provided in Algorithm \ref{alg:cerebellum}. After training, the network can operate without reference torques, relying solely on the learned synaptic weights to generate gravity compensation torque. 
\begin{algorithm}
	\caption{Training process of network for one motor pattern}
	\label{alg:cerebellum}
	\small
	\begin{algorithmic}[1]
		\STATE initialize load $M$, trajectory length $L$, batch size $bs$, learning rate $lr$, momentum $mo$, epochs $ep$, random noise $\bm{n}$, network loss $loss$
		\STATE initialize reference torque $\bm{\tau}_\text{ref}[1:L]$, network output torque $\bm{\tau}_{\text{cb}}[1:bs]$
		\STATE initialize optimizer $optim(lr,mo)$
		\STATE initialize network
		\STATE load training trajectory $\bm{q}_d[1:L]$
		\STATE load robotic arm urdf file and initialize dynamic model according to $M$
		
		\FOR {$t=1$ to $L$}
		\STATE read joint position $\bm{q}_d[t]$
		\STATE generate reference torque  $\bm{\tau}_\text{ref}(\bm{q}_d[t])$ using dynamic model
		\STATE update $\bm{\tau}_\text{ref}[t]$
		\ENDFOR
		
		\FOR {$i=1$ to $ep$}
		\FOR {$j=1$ to $L$ by $bs$}
		
		\STATE update random noise $\bm{n}$
		\STATE add noise to a batch of reference torque $\bm{\tau}_\text{ref,n}[j:j+bs]=\bm{\tau}_\text{ref}[j:j+bs]+\bm{n}$
		\STATE generate network output $\bm{\tau}_{\text{cb}}[j:j+bs]$
		\STATE calculate error between network output and reference torque $loss(\bm{\tau}_\text{ref,n}[j:j+bs],\bm{\tau}_{\text{cb}}[j:j+bs])$
		\STATE back propagate $loss$
		\STATE update network weights using $optim$
		
		\ENDFOR
		\ENDFOR 
		
	\end{algorithmic}
\end{algorithm} 

During back propagation, a surrogate gradient function is employed, given by  
\begin{equation}
	\sigma(x)=\frac{1}{1+\exp^{-\nu x}}
\end{equation}  
where $\nu$ is a scaling factor. In this study, horizontal circular and inclined circular trajectories are used for training, with their mathematical expressions defined as  
\begin{align}
	\text{horizontal circle:}&\begin{cases}
		x=x_0+R_c\cos(\frac{2\pi t}{T_c}) \\
		y=y_0+R_c\sin(\frac{2\pi t}{T_c}) \\
		z=z_0
	\end{cases}\\
	\text{inclined circle:}&\begin{cases}
		x=x_1+R_c\cos(\frac{2\pi t}{T_c})\cos\theta_0 \\
		y=y_1+R_c\sin(\frac{2\pi t}{T_c}) \\
		z=z_1+R_c\cos(\frac{2\pi t}{T_c})\sin\theta_0
	\end{cases}
\end{align}  
The parameters of neurons in each layer, as well as other settings used during training, are summarized in Table \ref{tab:cere_para}.
\begin{table}
	\centering
	\caption{Training and neuron parameters of the cerebellum module}
	\renewcommand\arraystretch{1.5}
	\begin{tabular}{cc|cc}
		\toprule
		Parameter  & Value & Parameter & Value       \\
		\midrule
		Epochs   & 50 each trajectory & \multirow{3}{*}{$u_{\text{reset}}$} & GC: 0.0\\
		Learning rate & 0.01 &   & PC: 0.0               \\
		Momentum&  0.5   &        & DCN: 0.0       \\
		\cline{3-4}
		Noise& $N(0,1)$ & \multirow{3}{*}{$u_{\text{fire}}$} & GC: 0.1 \\
		Batch size & 10 & & PC: 0.1  \\
		Optimizer &  SGD  & & DCN: $\infty$\\
		\cline{3-4}
		$\theta_0$ & $-\frac{\pi}{6}$  & \multirow{3}{*}{$\tau_\text{mb}$} & GC: 10.0 \\
		$R_c$ & 0.14 & & PC: 10.0\\
		$T_c$ & 3 & & DCN: 5.0\\
		\cline{3-4}
		$(x_0,y_0,z_0)$ & (0.54,0.0,0.45)  & $\nu$ & 5.0 \\
		$(x_1,y_1,z_1)$ & (0.63,-0.11,0.3)  & $T$  & 10\\
		\bottomrule
	\end{tabular}
	\label{tab:cere_para}
\end{table}

To realize motor patterns for different load masses, two networks, as depicted in Fig.~\ref{fig:cerebellum_train}, are trained separately in light pattern (with no load) and in heavy pattern (with 3 kg load) using the supervised learning procedure described in Algorithm~\ref{alg:cerebellum}. These trained networks are then combined to form the complete cerebellum module, as illustrated in Fig.~\ref{fig:cerebellum_predict}.

In summary, after training, the cerebellum module receives as input the actual joint positions of the robotic arm, $\bm{q} \in \mathbb{R}^7$, and outputs a gravity compensation matrix composed of two torque vectors, i.e., $\bm{T}_{\text{cb}}=(\bm{\tau}_\text{cb,L},\bm{\tau}_\text{cb,H})$, $\bm{\tau}_\text{cb,L} \in \mathbb{R}^7$ and $\bm{\tau}_\text{cb,H} \in \mathbb{R}^7$. This matrix corresponds to the two learned patterns of the cerebellum module and provides gravity compensation for trajectory tracking tasks under both light and heavy pattern. During trajectory tracking tasks, the prediction process of the cerebellum module is described in Algorithm \ref{alg:cerebellum1}, which can be implemented as Sub-thread 4 in Algorithm \ref{alg:cbmc}.
\begin{algorithm}
	\caption{Prediction process of the cerebellum module}
	\label{alg:cerebellum1}
	\small
	\begin{algorithmic}[1]
		\STATE load the cerebellum module network
		\STATE initialize gravity compensation matrix $\bm{T}_{\text{cb}}$
		\STATE read joint position $\bm{q}$
		\STATE feed $\bm{q}$ into the cerebellum module network, and generate two output $\bm{\tau}_{\text{cb,L}},\bm{\tau}_{\text{cb,H}}$
		\STATE update $\bm{T}_{\text{cb}}=(\bm{\tau}_{\text{cb,L}},\bm{\tau}_{\text{cb,H}})$
		
	\end{algorithmic}
\end{algorithm}

\subsection{Thalamus module}
As described in the previous sub-section, although we trained two neural networks to form the cerebellum module—thereby generating gravity compensation torques corresponding to two specific load masses—the cerebellum module alone remains insufficient for handling arbitrary load. To address this limitation, we developed the thalamus module. By utilizing torque sensor data from the robotic arm’s end-effector, the thalamus module network is able to “sense” the mass of the object, akin to human perception, and accordingly assign adaptive weights to the two gravity compensation torques produced by the cerebellum module’s motor patterns via reinforcement learning.

\begin{figure}[h]
	\centering
	\includegraphics[width=0.7\linewidth]{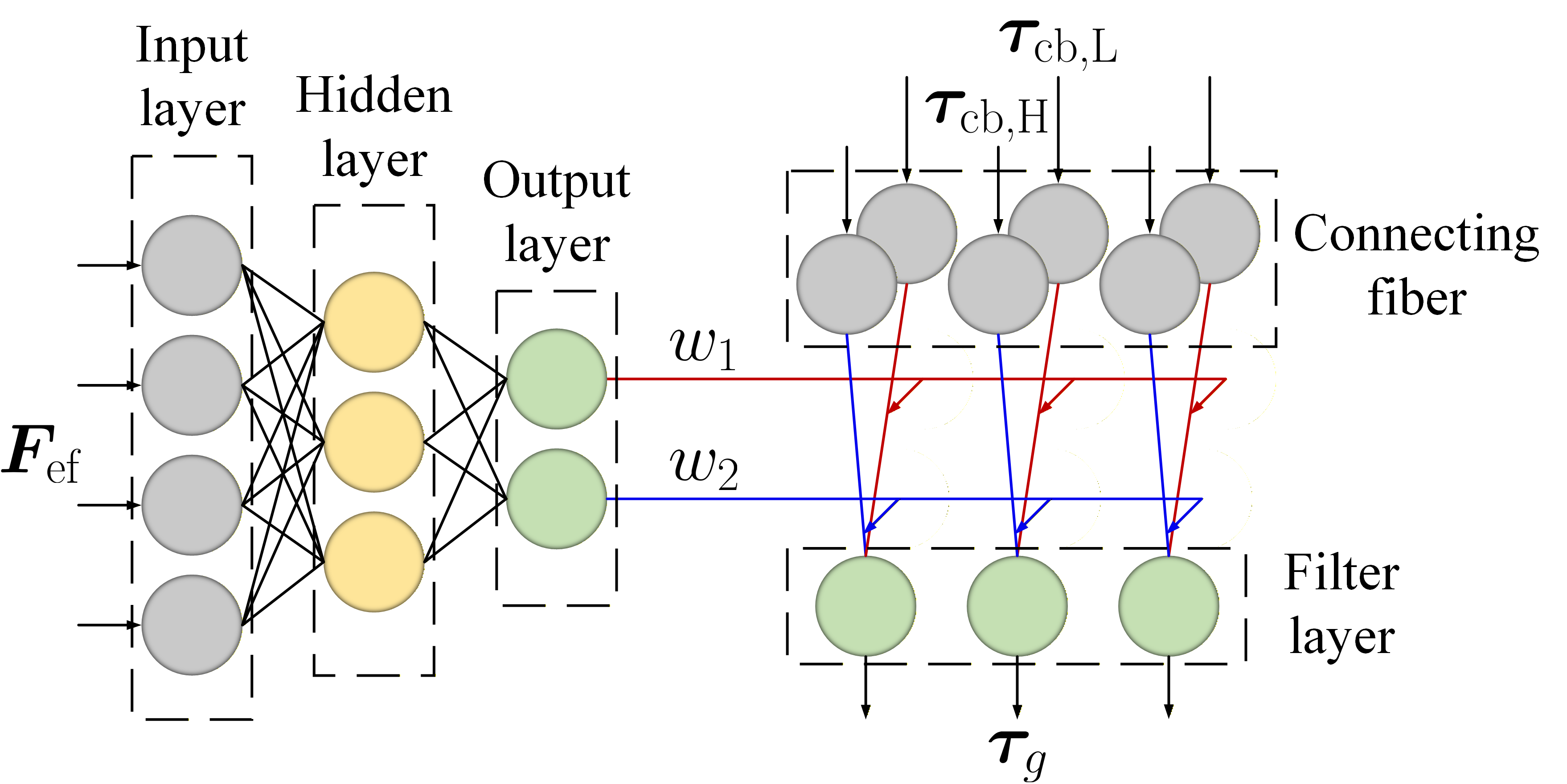}
	\caption{Structure of the thalamus module.}
	\label{fig:thalamus}
\end{figure}

The thalamus module contains five layers of neurons, as illustrated in Fig.~\ref{fig:thalamus}. Its input $\bm{F}_\text{ef}\in\mathbb{R}^6$ and $\bm{T}_{\text{cb}}=(\bm{\tau}_\text{cb,L},\bm{\tau}_\text{cb,H})\in\mathbb{R}^{7\times 2}$, correspond to the six-dimensional force/torque sensor data measured at the end-effector, and the cerebellum module output, respectively. Its output, $\bm{\tau}_g\in\mathbb{R}^7$, represents the gravity compensation torque for arbitrary load mass within a certain range. 

The input layer is fully connected to a hidden layer consisting of five LIF neurons, which in turn connects to an output layer of two non-spiking LIF neurons. The membrane potential of the output neurons are then passed through a Softmax function to produce the two weight coefficients. The network is trained via reinforcement learning, and the loss function designed for this network comprises two terms, minimizing both the joint position error and the joint velocity error, i.e.,
\begin{equation}
	\text{Loss}=\sum_{i=1}^{7}100*(q_{d,i}-q_i)+\sum_{i=1}^{7}10^{-4}\cdot(\dot{q}_{d,i}-\dot{q}_i)
\end{equation}  
Similar to the cerebellum module, the thalamus module adopts rate-based encoding with a time window length of $T$, and employs the sigmoid function as the surrogate gradient function.
\begin{table}
	\centering
	\caption{Training and neuron parameters of the thalamus module}
	\renewcommand\arraystretch{1.5}
	\begin{tabular}{cc|cc}
		\toprule
		Parameter  & Value & Parameter & Value       \\
		\midrule
		\multirow{2}{*}{Learning rate}   & \multirow{2}{*}{0.01} & $\nu$ & 5.0 \\
		& & $T$ & 10\\
		\cline{3-4}
		Weight decay & 0.001 & \multirow{3}{*}{$u_{\text{reset}}$}  & Hidden layer: 0.0               \\
		Optimizer &  SGD   &        & Output layer: 10.0         \\
		Momentum &  0.5   &        & Filter layer: 0.0       \\
		\hline
		\multirow{3}{*}{$\tau_\text{mb}$} & Hidden layer: 5.0  & \multirow{3}{*}{$u_{\text{fire}}$} & Hidden layer: 0.01 \\
		&  Output layer: 5.0   & & Output layer: $\infty$  \\
		&  Filter layer: 5.0 & & Filter layer: $\infty$ \\
		\bottomrule
	\end{tabular}
	\label{tab:thala_para}
\end{table}

After obtaining $\bm{w}$ from the output layer, it is then combined with data from the connecting fiber (linking the thalamus module and the cerebellum module) to generate a weighted output, yielding
\begin{equation}
	\begin{split}
		\delta\bm{u}_\text{ft}(t) &= (\delta u_\text{ft,1}(t),\delta u_\text{ft,2}(t),\cdots,\delta u_\text{ft,7}(t))^\top\\
		&= (\bm{\tau}_\text{cb,L}, \bm{\tau}_\text{cb,H})\cdot\bm{w}=w_1\bm{\tau}_\text{cb,L}+w_2\bm{\tau}_\text{cb,H}
	\end{split}
\end{equation} 
This would then be injected as an increment to the membrane potentials of seven non-spiking LIF neurons that serve as a filter layer. Specifically, the membrane potential of each filtering-layer neuron at time $t$ is given by 
\begin{equation}
	\begin{aligned}
		u_{\text{ft},i}(t)&=u_{\text{ft},i}(t-1)-\frac{u_{\text{ft},i}(t-1)-u_{\text{reset}}}{\tau_\text{mb}}\\
		&+\frac{\delta u_{\text{ft},i}(t)}{\tau_\text{mb}},i=1,\cdots,7
	\end{aligned}
\end{equation} 

Finally we take the membrane potential of each filter-layer neuron at time $t$ as the gravity compensation torque, i.e., $\bm{\tau}_g=\bm{u}_\text{ft}(t)=(u_{\text{ft},1}(t),u_{\text{ft},2}(t),\cdots,u_{\text{ft},7}(t))^\top$. The parameters of the neurons in each layer of this network, as well as the training parameters, are summarized in Table \ref{tab:thala_para}. During trajectory tracking tasks, the learning procedure of the thalamus module follows Algorithm \ref{alg:thalamus}, which can be directly implemented as Sub-thread 3 in Algorithm \ref{alg:cbmc}.
\begin{algorithm}
	\caption{Training process of the thalamus module}
	\label{alg:thalamus}
	\small
	\begin{algorithmic}[1]
		\STATE initialize gravity compensation torque $\bm{\tau}_g$, weight coefficients $\bm{w}$, learning rate $lr$, momentum $mo$, network loss $loss$, optimizer $optim(lr,mo)$
		\STATE read $\bm{T}_{{\text{cb}}},\bm{F}_\text{ef}$
		\STATE feed $\bm{F}_\text{ef}$ into the network and generate weight coefficients $\bm{w}$
		\STATE calculate $\bm{T}_{\text{cb}}\cdot\bm{w}$ and feed the result into the filter layer neurons
		\STATE read the membrane potential of the filter layer $\bm{u}_\text{ft}$, update $\bm{\tau}_g$
		\STATE read joint position and velocity data  $\bm{\dot{q}}_d,\bm{q}_d,\bm{\dot{q}},\bm{q}$
		\STATE calculate and update $loss$
		\STATE back propagate $loss$
		\STATE update network weights using $optim$
		
	\end{algorithmic}
\end{algorithm}

\subsection{Brainstem module}
As stated in section \ref{sec:bs}, the brainstem module is to provide appropriate feedback control coefficients during operation to compensate for the nonlinear dynamics of the robotic arm and the coupling effects among joints. Its network is implemented as a three-layer fully connected SNN, as illustrated in Fig.~\ref{fig:framework3}. The input variable $[\bm{q}_d,\bm{\dot{q}}_d,\bm{q},\bm{\dot{q}}]\in\mathbb{R}^{28}$ is a concatenated vector consisting of the desired joint positions and velocities, as well as the actual joint positions and velocities. The output $\bm{K}_P = diag(k_{p,1},k_{p,2},\cdots,k_{p,7})$, $\bm{K}_V = diag(k_{v,1},k_{v,2},\cdots,k_{v,7})$ corresponds to the proportional and derivative coefficients for each of the seven joints.

The input vector is fully connected to a hidden layer of 10 LIF neurons, which in turn is fully connected to an output layer of 14 non-spiking LIF neurons. The network is trained using reinforcement learning, with the loss function defined as 
\begin{equation}
	\begin{aligned}
		\text{Loss}&=\sum_{i=1}^{7}100*(q_{d,i}-q_i)+\sum_{i=1}^{7}10^{-4}\cdot(\dot{q}_{d,i}-\dot{q}_i)\\
		&+\sum_{i=1}^{7}10^{-3}\cdot\tau_i^2
	\end{aligned}
\end{equation} 
In this formulation, the first term and second term minimize position and tracking error to ensure accurate trajectory tracking, while the third term penalizes torque usage to reduce joint energy consumption.

Similar to the cerebellum module, the brainstem module adopts rate–based encoding with a time window length of $T$, and the surrogate gradient function is chosen as the sigmoid function.

After decoding through the non-spiking LIF neurons in the output layer, we obtain a set of membrane potentials $\bm{u}_\text{bs}=(u_{\text{bs},1},\cdots,u_{\text{bs},14})\in\mathbb{R}^{14}$. By applying a scaling transformation, these membrane potentials can be mapped to the proportional and derivative coefficients, i.e.,
\begin{align}
	\begin{split}
		\bm{K}_P &= diag(k_{p,1},k_{p,2},\cdots,k_{p,7})\\
		&= a\cdot diag(u_\text{bs,1},u_\text{bs,2},\cdots,u_\text{bs,7})
	\end{split}
	\\
	\begin{split}
		\bm{K}_V &= diag(k_{v,1},k_{v,2},\cdots,k_{v,7})\\
		&= b\cdot diag(u_\text{bs,8},u_\text{bs,9},\cdots,u_\text{bs,14})
	\end{split}
\end{align} 
where $a$ and $b$ are two scaling parameters. The parameters of the neurons in each layer of the network, as well as the parameters used during training, are summarized in Table \ref{tab:brainstem_para}.
\begin{table}
	\centering
	\caption{Training and neuron parameters of the brainstem module}
	\renewcommand\arraystretch{1.5}
	\begin{tabular}{cc|cc}
		\toprule
		Parameter  & Value & Parameter & Value       \\
		\midrule
		Learning Rate   & 0.01 & \multirow{2}{*}{$u_{\text{reset}}$} & Hidden layer: 0.0\\
		Momentum & 0.5 &   & Output layer: 10.0               \\
		\cline{3-4}
		Optimizer &  SGD   &    \multirow{2}{*}{$u_{\text{fire}}$}    & Hidden layer: 0.005       \\
		$\nu$ & 5.0  &  & Output layer: $\infty$ \\
		\cline{3-4}
		(a,b) & (20,2)  & \multirow{2}{*}{$\tau_\text{mb}$} & Hidden layer: 5.0  \\
		$T$ &  15  & & Output layer: 5.0\\
		\bottomrule
	\end{tabular}
	\label{tab:brainstem_para}
\end{table}

During trajectory tracking tasks, the learning process of the brainstem module is described in Algorithm~\ref{alg:brainstem}, which can be implemented as Sub-thread 2 within Algorithm\ref{alg:cbmc}.
\begin{algorithm}
	\caption{Training process of the brainstem module}
	\label{alg:brainstem}
	\small
	\begin{algorithmic}[1]
		\STATE initialize coefficients $\bm{K}_V,\bm{K}_P$, learning rate $lr$, momentum $mo$, network loss $loss$, optimizer $optim(lr,mo)$
		\STATE read joint position and velocity data $\bm{\dot{q}}_d,\bm{q}_d,\bm{\dot{q}},\bm{q}$
		\STATE feed $[\bm{q}_d,\bm{\dot{q}}_d,\bm{q},\bm{\dot{q}}]$ into the network
		\STATE read the potential of the output layer neurons $\bm{u}_\text{bs}$ and scale them through scaling parameters
		\STATE update $\bm{K}_V,\bm{K}_P$
		\STATE read joint position and velocity data $\bm{\dot{q}}_d,\bm{q}_d,\bm{\dot{q}},\bm{q}$
		\STATE calculate and update $loss$
		\STATE back propagate $loss$
		\STATE update network weights using $optim$
		
	\end{algorithmic}
\end{algorithm}

\subsection{Spinal cord module}
According to section \ref{sec:sc}, the spinal cord module is designed to integrate proprioceptive information and the outputs of the aforementioned modules to generate real-time motor control commands. For high-performance robotic arms, torque control typically requires the control algorithm to achieve frequencies of 1 kHz or even higher. Consequently, the spinal cord module must exhibit very high computational efficiency.

The structure of this module is illustrated in Fig.~\ref{fig:framework3}. The input variables $[\bm{q}_d,\bm{\dot{q}}_d,\bm{q},\bm{\dot{q}}]\in\mathbb{R}^{28}$ consist of the planned positions and velocities as well as the actual positions and velocities of seven joints, while the output $\bm{\tau}\in\mathbb{R}^{7}$ represents the control torque for each joint. After entering the spinal cord module, the planned trajectory and proprioceptive feedback are first encoded into discrete spike trains, with each joint being processed independently. Specifically, the planned position and velocity signals $q_{d,i},\dot{q}_{d,i},i=1,\cdots,7$ excite the input fibers, whereas the actual proprioceptive feedback $q_{i},\dot{q}_{i},i=1,\cdots,7$ inhibit the input fibers. Mathematically, this process is equivalent to computing their difference.

The outputs of the input fibers are subsequently passed into the Poisson encoding stage, given by 
\begin{align}
	s_{q,i,j}=\begin{cases}
		0, &-u_j<\frac{q_{d,i}-q_{i}}{\Delta q_i}<u_j,j=1,\cdots,k\\
		1, &\frac{q_{d,i}-q_{i}}{\Delta q_i}\geq u_j,j=1,\cdots,\frac{k}{2}\\
		-1, &-\frac{q_{d,i}-q_{i}}{\Delta q_i}\geq u_j,j=\frac{k}{2},\cdots,k
	\end{cases}\\
	s_{\dot{q},i,j}=\begin{cases}
		0, &-u_j<\frac{\dot{q}_{d,i}-\dot{q}_{i}}{\Delta \dot{q}_i}<u_j,j=1,\cdots,k\\
		1, &\frac{\dot{q}_{d,i}-\dot{q}_{i}}{\Delta \dot{q}_i}\geq u_j,j=1,\cdots,\frac{k}{2}\\
		-1, &-\frac{\dot{q}_{d,i}-\dot{q}_{i}}{\Delta \dot{q}_i}\geq u_j,j=\frac{k}{2},\cdots,k
	\end{cases}
\end{align} 
Here, $u_j\sim U(0,1),j=1,\cdots,k$ denotes $k$ independent uniformly distributed random variables, ${s_{q,i,j}}, {s_{\dot{q},i,j}},i=1,\cdots,7,j=1,\cdots,k$ denote the spike trains encoding the position and velocity information of each joint at each time step, respectively. The terms $\Delta q_i$ and $\Delta \dot{q}_i, i=1,\cdots,7$ correspond to the maximum position and velocity errors for each joint, respectively.

From an intuitive perspective, each input fiber is connected to a group of $k$ neurons, where the first half $\frac{k}{2}$ are responsible for processing positive inputs and the second half $\frac{k}{2}$ handle negative inputs. This structure corresponds to the agonist–antagonist pairing characteristic of human muscles. Each neuron generates a uniformly distributed random number in the range of $[0,1]$. When the output of the input fiber is positive, the first $\frac{k}{2}$ neurons are activated: if the absolute value of each neuron's input exceeds the random number, the encoding result is set to $1$; otherwise, it is $0$. Conversely, when the output of the input fiber is negative, the second $\frac{k}{2}$ neurons are activated: if the absolute value of each neuron's input exceeds the random number, the encoding result is set to $-1$; otherwise, it is $0$. The encoding effect is influenced by three parameters: $\Delta q_i$, $\Delta \dot{q}_i$, $i=1,\cdots,7$, and $k$. Smaller values of $\Delta q_i$ and $\Delta \dot{q}_i$ lead to higher encoding precision, but reduce the dynamic range; a larger $k$ increases encoding precision but reduces computational efficiency.

Following this encoding process, two sets of spike trains are generated—one representing position information and the other representing velocity information. These spike trains are then transmitted to stimulate the activity of the non-spiking LIF neuron for each joint, with stimulation strength modulated by the feedback control coefficients $k_{p,i}$ and $k_{v,i}$, $i=1,\cdots,7$ provided by the brainstem module. Accordingly, at each time step, the membrane potential increment of the output non-spiking LIF neurons can be expressed as 
\begin{equation}
	\begin{aligned}
		\delta u_\text{sp,i}(t)&=\tau_{g,i}+k_{p,i}\cdot\sum_{j=1}^{100}s_{q,i,j}\\
		&+k_{v,i}\cdot\sum_{j=1}^{100}s_{\dot{q},i,j},i=1,\cdots,7
	\end{aligned}
\end{equation} 
Here, $\bm{\tau}_g=(\tau_{g,1},\tau_{g,2},\cdots,\tau_{g,7})^\top$ denotes the gravity compensation torque generated by the thalamus module. Consequently, the membrane potentials at each time step can be written as 
\begin{equation}
	\begin{aligned}
		u_{\text{sp},i}(t)&=u_{\text{sp},i}(t-1)-\frac{u_{\text{sp},i}(t-1)-u_{\text{reset}}}{\tau_\text{mb}}\\
		&+\frac{\delta u_{\text{sp},i}(t)}{\tau_\text{mb}},i=1,\cdots,7
	\end{aligned}
\end{equation} 

Finally, similar to the thalamus module, we take $\bm{\tau}=\bm{u}_\text{sp}=(u_{\text{sp},1}(t),u_{\text{sp},2}(t),\cdots,u_{\text{sp},7}(t))^\top$ as the control torque to be sent in real time to the actuators. The neuronal and encoding parameters of the spinal cord module are summarized in Table~\ref{tab:spinal_para}.
\begin{table}
	\centering
	\caption{Neuron parameters of the spinal cord module}
	\renewcommand\arraystretch{1.5}
	\begin{tabular}{cc|cc}
		\toprule
		Parameter  & Value & Parameter  & Value \\
		\midrule
		$u_{\text{reset}}$ & 0.0 & $\Delta q_i,i=1,\cdots,7$ & 0.5 \\
		$u_{\text{fire}}$ & $\infty$ & $\Delta \dot{q}_i,i=1,\cdots,7$ & 0.5 \\
		$\tau_\text{mb}$ & 2.0 & k & 100 \\
		\bottomrule
	\end{tabular}
	\label{tab:spinal_para}
\end{table}

During trajectory tracking tasks, the operational process of the spinal cord module is illustrated in Algorithm~\ref{alg:spinal}, which can be implemented as Sub-thread 1 within Algorithm\ref{alg:cbmc}.
\begin{algorithm}
	\caption{Prediction process of the spinal cord module}
	\label{alg:spinal}
	\small
	\begin{algorithmic}[1]
		\STATE initialize control torque $\bm{\tau}$
		\STATE read joint position and velocity data  $\bm{\dot{q}}_d,\bm{q}_d,\bm{\dot{q}},\bm{q}$
		\STATE read proportional and derivative coefficients $\bm{K}_V,\bm{K}_P$ and gravity compensation torque  $\bm{\tau}_g$
		\STATE poisson encode $\bm{\dot{q}}_d,\bm{q}_d,\bm{\dot{q}},\bm{q}$ to generate two sets of spike trains $[s_{q,i,j}],[s_{\dot{q},i,j}]$
		\STATE modulate the spike trains with $\bm{K}_V,\bm{K}_P$
		\STATE feed $\bm{\tau}_g$ and the modulated spike trains into the non-spiking LIF neurons
		\STATE read potential and update $\bm{\tau}$
		
	\end{algorithmic}
\end{algorithm}

\section{Simulation results}\label{sec:simulation}

In this section, we conduct simulation experiments to validate that the module designs proposed in Section~\ref{sec:module} satisfy the requirements outlined in Section~\ref{sec:framework}, and to evaluate the resilience of the proposed framework.

The framework is implemented in Python (version 3.12.4), and all modules are constructed using PyTorch~\cite{paszke2019pytorch} (version 2.4.0). The spiking neurons involved in the framework are implemented with the SNN learning framework SpikingJelly~\cite{doi:10.1126/sciadv.adi1480} (version 0.0.0.0.14). The simulation environment employed is PyBullet~\cite{coumans2021} (version 3.2.6), an open-source robotics simulation tool whose underlying physics engine is based on the open-source Bullet Physics SDK. The controlled robotic arm is a Flexiv Rizon 4s consisting of seven actuated joints, seven links and a six-dimensional torque sensor mounted at the flange of the end-effector. The framework runs on a multi-core CPU, and the control cycle of the first-order, second-order and third-order level is set as $T_1=1$ms, $T_2=10$ms, $T_3=50$ms, respectively. The simulation time step is set to 1 ms, and the control frequency is configured to 1 kHz.

In all simulation experiments presented in this section, trajectory tracking accuracy is evaluated using the root mean square error (RMSE) of the joint positions. The RMSE at each time step is defined as  
\begin{equation}
	\text{RMSE}(t)=\sqrt{\sum_{i=1}^{7}(q_{d,i}(t)-q_i(t))^2}
\end{equation}
while the average RMSE over an entire trajectory is defined as \begin{equation}
	\text{RMSE}=\frac{1}{L}\sum_{t=1}^{L}\text{RMSE}(t)
\end{equation} 
where $L$ is the trajectory length. Three types of trajectories are employed in the simulations: horizontal circle, inclined circle, and spatial figure-eight. The first two trajectories have already been defined in section \ref{cbm}, while the third trajectory is defined as \begin{align}
	\text{spatial figure-eight:}&\begin{cases}
		x=x_0 + 0.5 R_e\sin(\frac{4\pi t}{T_e})\\
		y=y_0 + R_e\cos(\frac{2\pi t}{T_e}) \\
		z=z_0 + 0.08\sin(\frac{2\pi t}{T_e})
	\end{cases}
\end{align} 
where $x_0=0.61, y_0=0, z_0=0.3, R_e=0.14, T_e=3$.

\subsection{Module design validation}\label{sm1}
\begin{figure}
	\centering
	\subfloat[]{\includegraphics[width=0.48\linewidth]{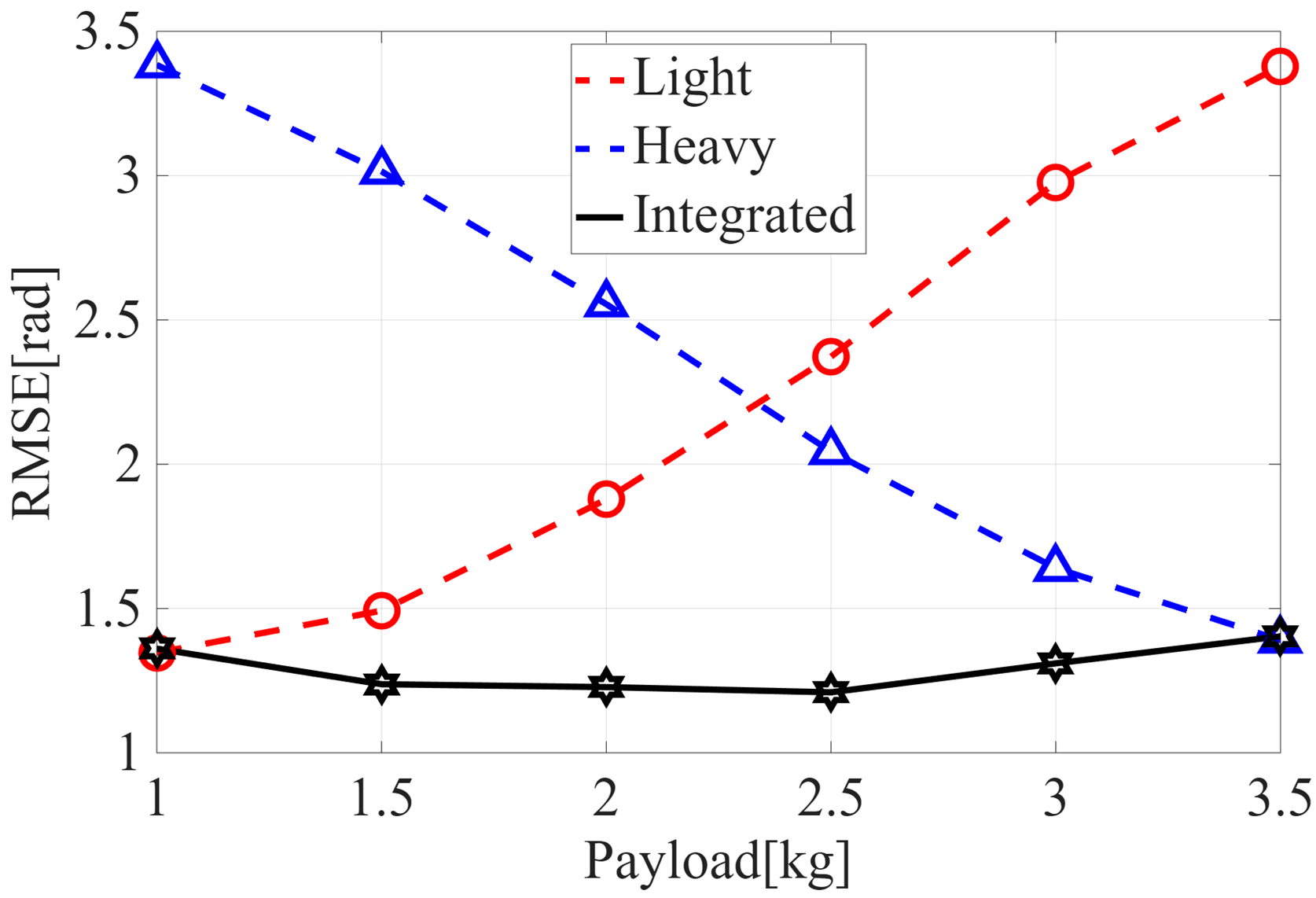}%
		\label{fig:thalamus1}}
	\hfil
	\subfloat[]{\includegraphics[width=0.48\linewidth]{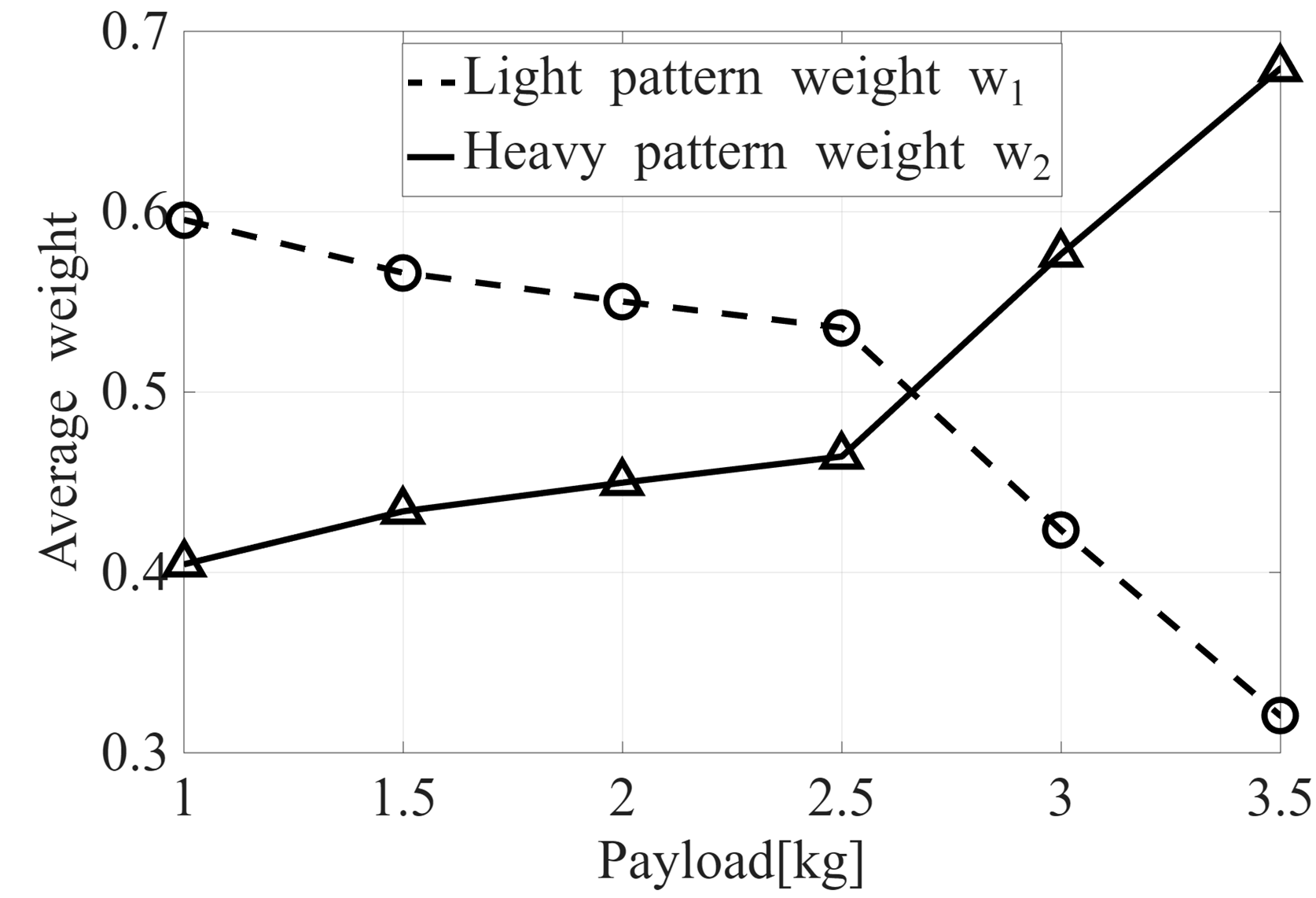}%
		\label{fig:thalamus2}}
	\caption{Performance of the cerebellum and thalamus modules. (a) Average RMSE across three trajectories using single motor pattern and two motor patterns. (b) Average motor pattern weight coefficients across three trajectories generated by the thalamus module.}
	\label{fig:thalamus result}
\end{figure}

Fig.~\ref{fig:thalamus result} illustrates the performance of the cerebellum and thalamus modules. As shown in Fig.~\ref{fig:thalamus1}, when only the light motor pattern is used, the tracking error increases with increasing load mass, whereas when only the heavy motor pattern is used, the tracking error decreases as the load mass increases. This indicates that the cerebellum module effectively encodes two distinct motor patterns: the light pattern corresponds to a load of approximately 1 kg, while the heavy pattern corresponds to a load of approximately 3.5 kg. However, neither pattern alone can accommodate arbitrary load masses within the range of 1 kg to 3.5 kg.

Owing to its online learning capability, the thalamus module can adaptively adjust the weighting coefficients assigned to these two motor patterns. As illustrated in Fig.~\ref{fig:thalamus2}, the thalamus module assigns a smaller weight to the light pattern and a larger weight to the heavy pattern as the load mass increases. Consequently, the weighted combination of the two patterns achieves consistently good performance across the entire load range from 1 kg to 3.5 kg. This effect is evident in Fig.~\ref{fig:thalamus1}, where the black curve remains below the red and blue curves.

This adaptive mechanism mirrors human motor control, in which movements are generated through linear combinations of motor primitives \cite{sanger2000human}, thereby highlighting the biomimetic nature of the cerebellum and thalamus modules. Overall, these results demonstrate that the designs of both modules satisfy the proposed requirements.

\begin{figure}
	\centering
	\includegraphics[width=\linewidth]{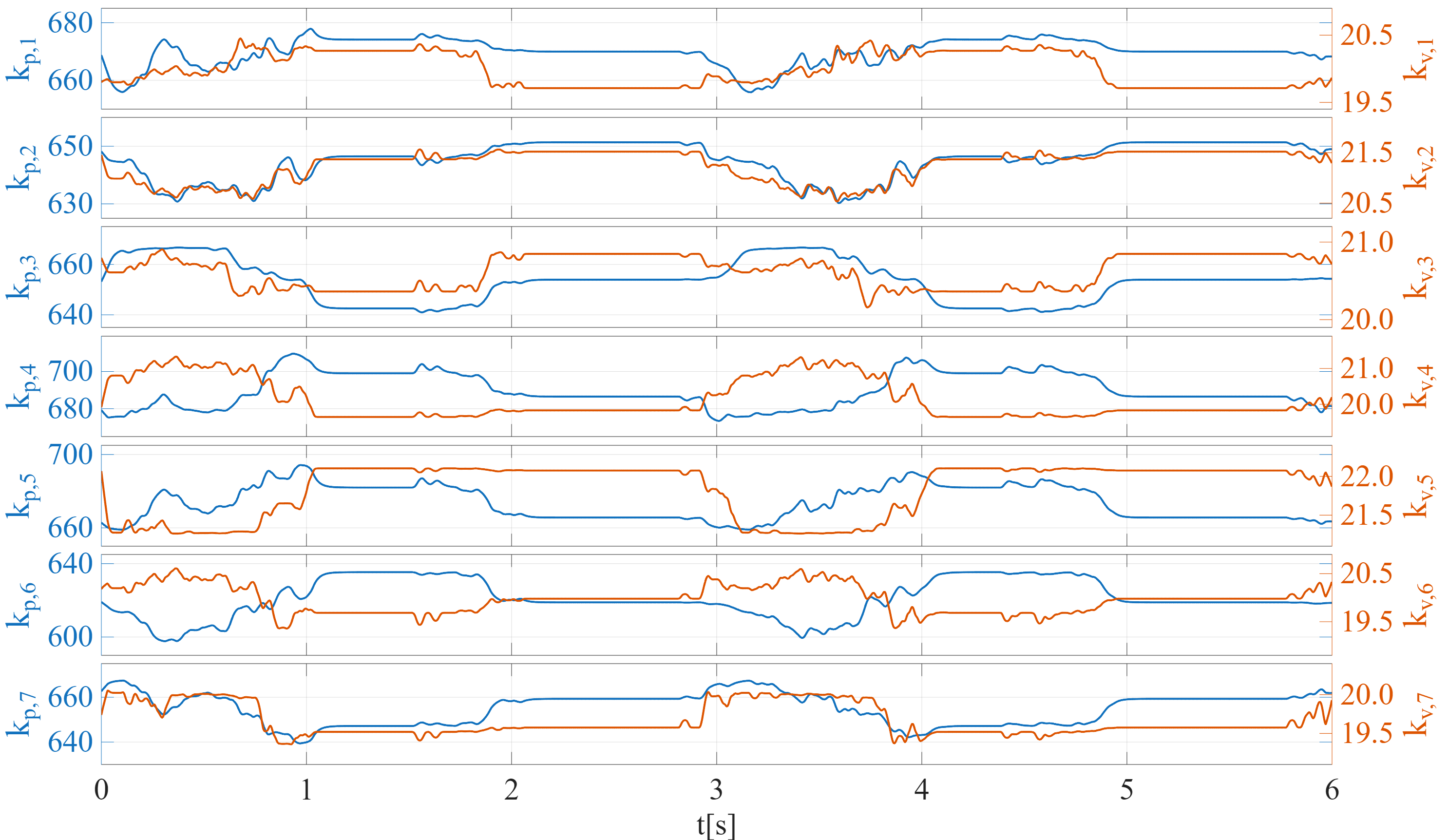}
	\caption{Performance of the brainstem module.}
	\label{fig:kpkd}
\end{figure}

Fig.~\ref{fig:kpkd} illustrates the evolution of the average stiffness coefficients $k_{p, i}, i=1,\cdots,7$ and damping coefficients $k_{v, i}, i=1,\cdots,7$ for each joint across three trajectories. The results show that the brainstem module can adaptively generate feedback control gains for the spinal cord module. Notably, the stiffness and damping coefficients of joints 4 and 6 generally exhibit opposite trends. Quantitative analysis confirms this observation: the correlation coefficients between $k_{p, 4}$ and $k_{v, 4}$, and between $k_{p, 6}$ and $k_{v, 6}$, are $-0.5640$ and $-0.8515$, respectively. This behavior is consistent with human motor control. Physiological studies have shown that joint stiffness and damping in the human arm are inversely correlated \cite{bennett1992time}, indicating that the brainstem module exhibits biomimetic characteristics. Overall, these results demonstrate that the design of the brainstem module satisfies the proposed requirements.

\subsection{Resilience performance evaluation}
\begin{figure*}
	\begin{minipage}[b]{\linewidth}
		\centering
		\subfloat[]{\includegraphics[width=0.5\linewidth]{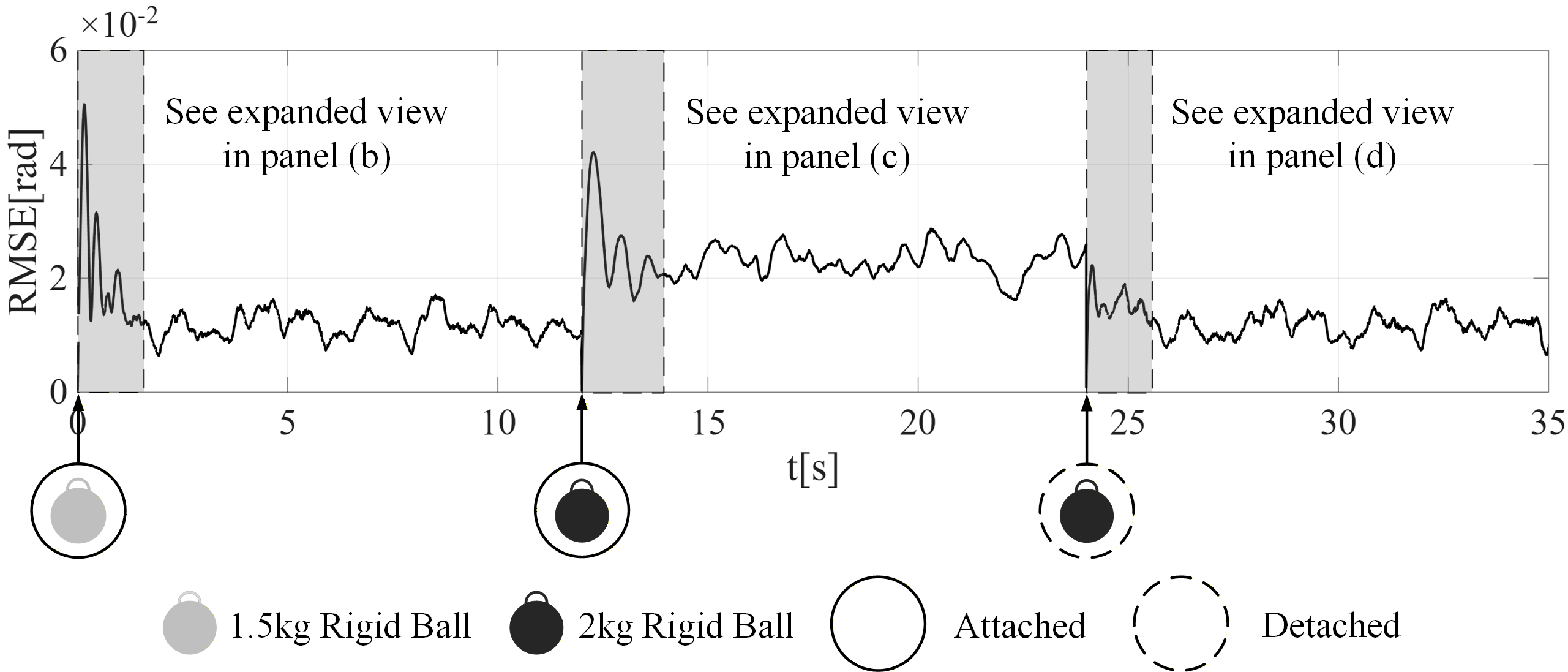}%
			\label{fig:vary_payload_error-a}}
		\subfloat[]{\includegraphics[width=0.245\linewidth]{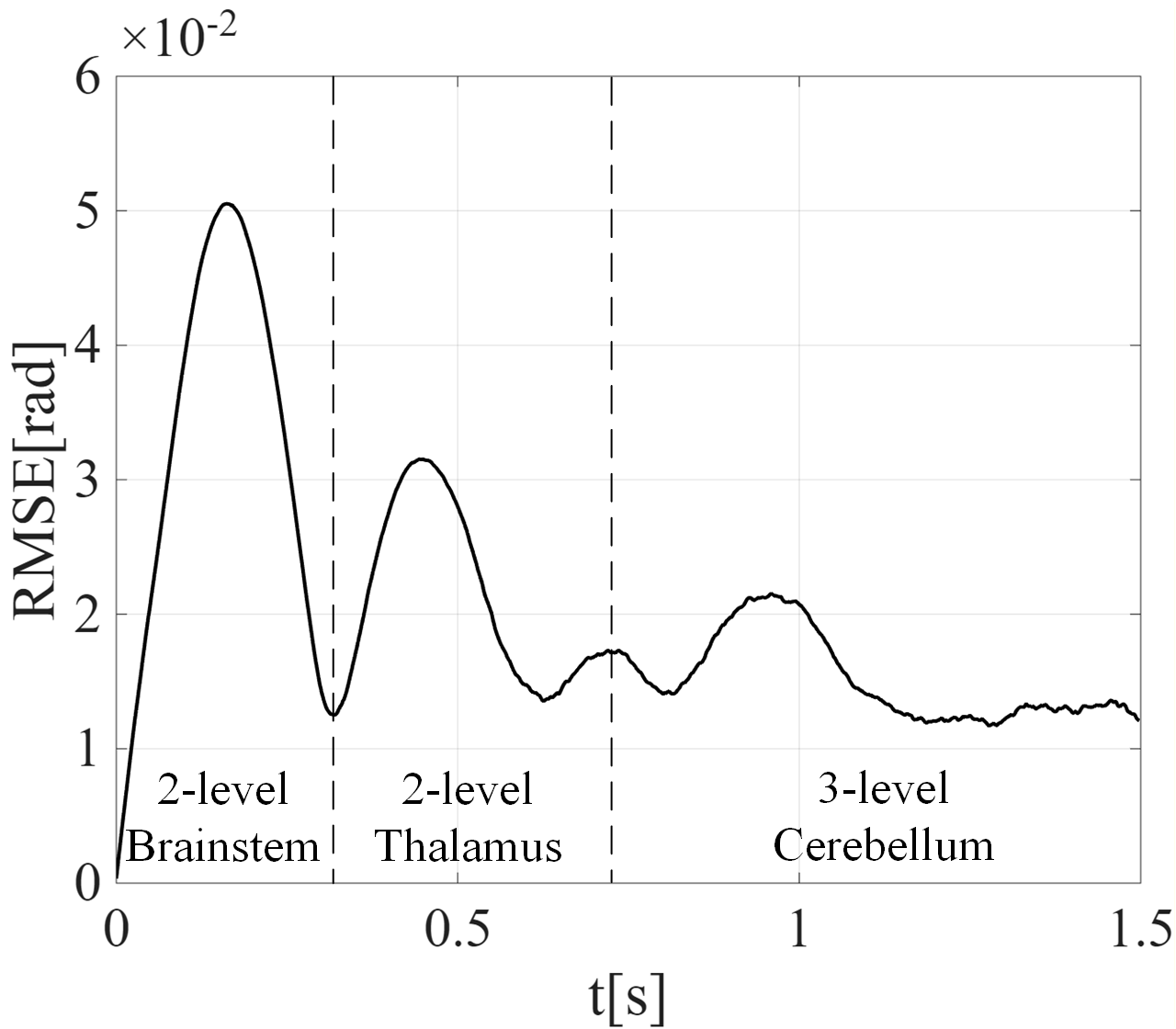}%
			\label{fig:vary_payload_error-b}}
		\subfloat[]{\includegraphics[width=0.245\linewidth]{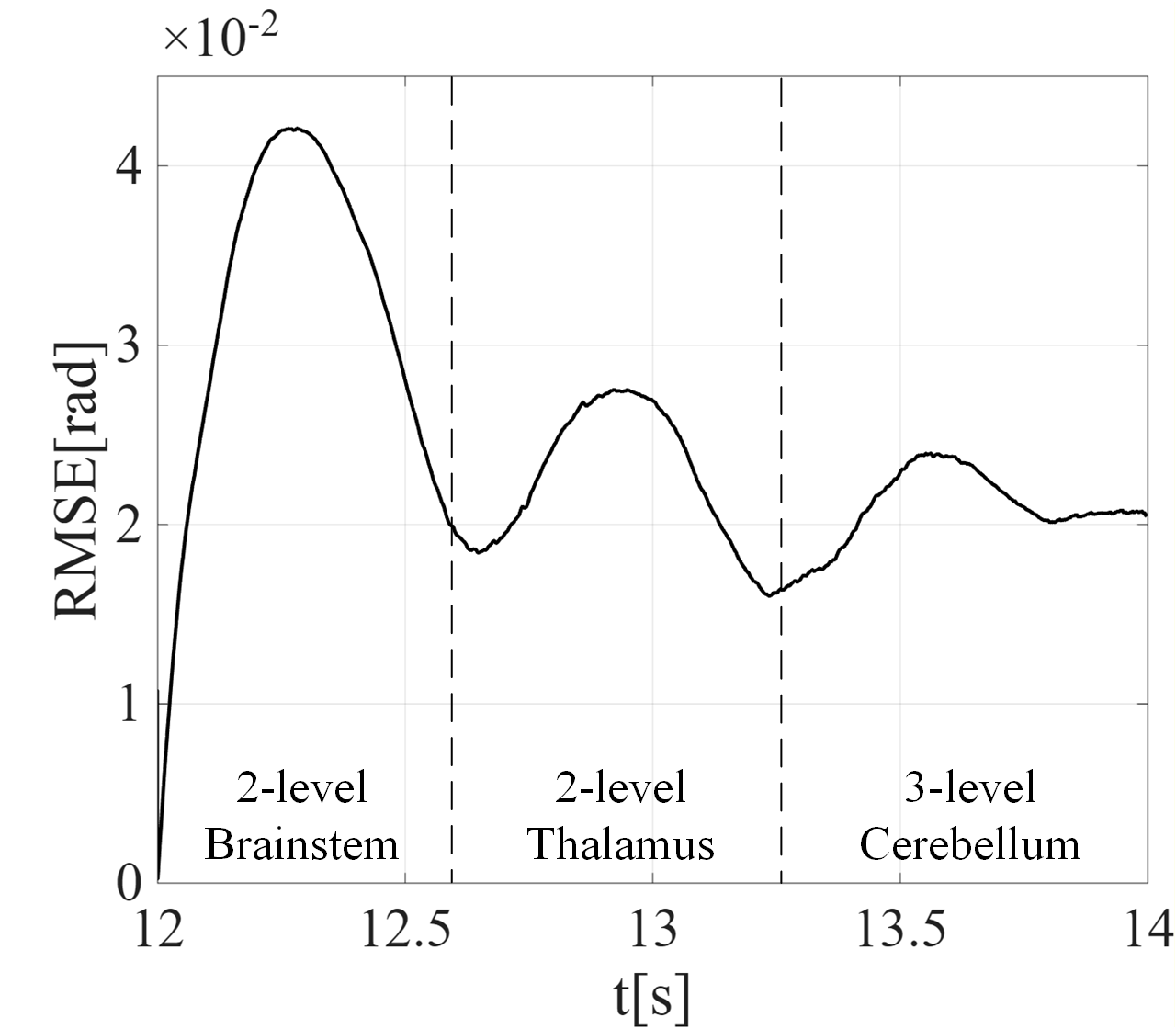}%
			\label{fig:vary_payload_error-c}}
	\end{minipage}
	\hfil
	\begin{minipage}[b]{\linewidth}
		\centering
		\subfloat[]{\includegraphics[width=0.25\linewidth]{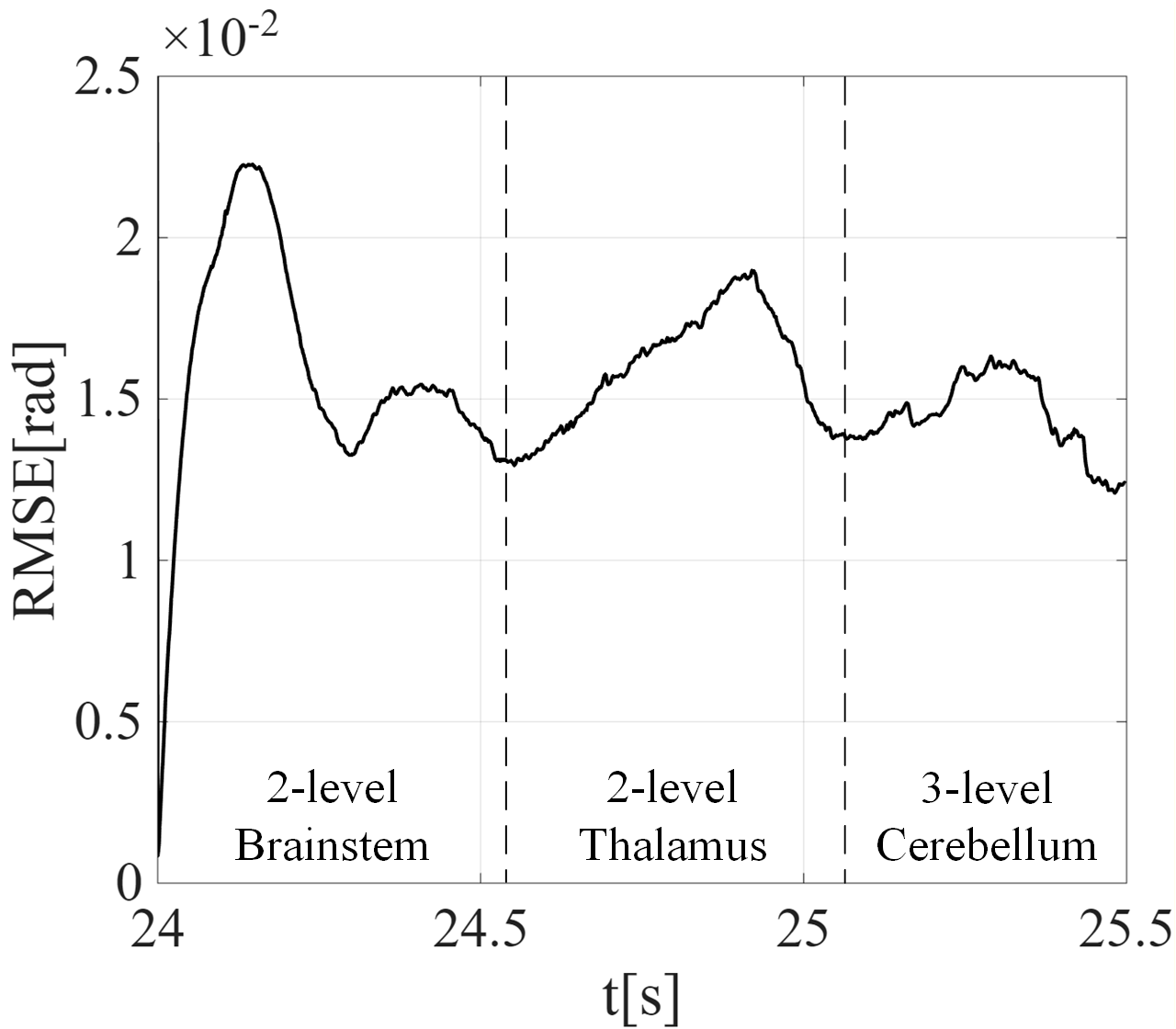}%
			\label{fig:vary_payload_error-d}}
		\subfloat[]{\includegraphics[width=0.25\linewidth]{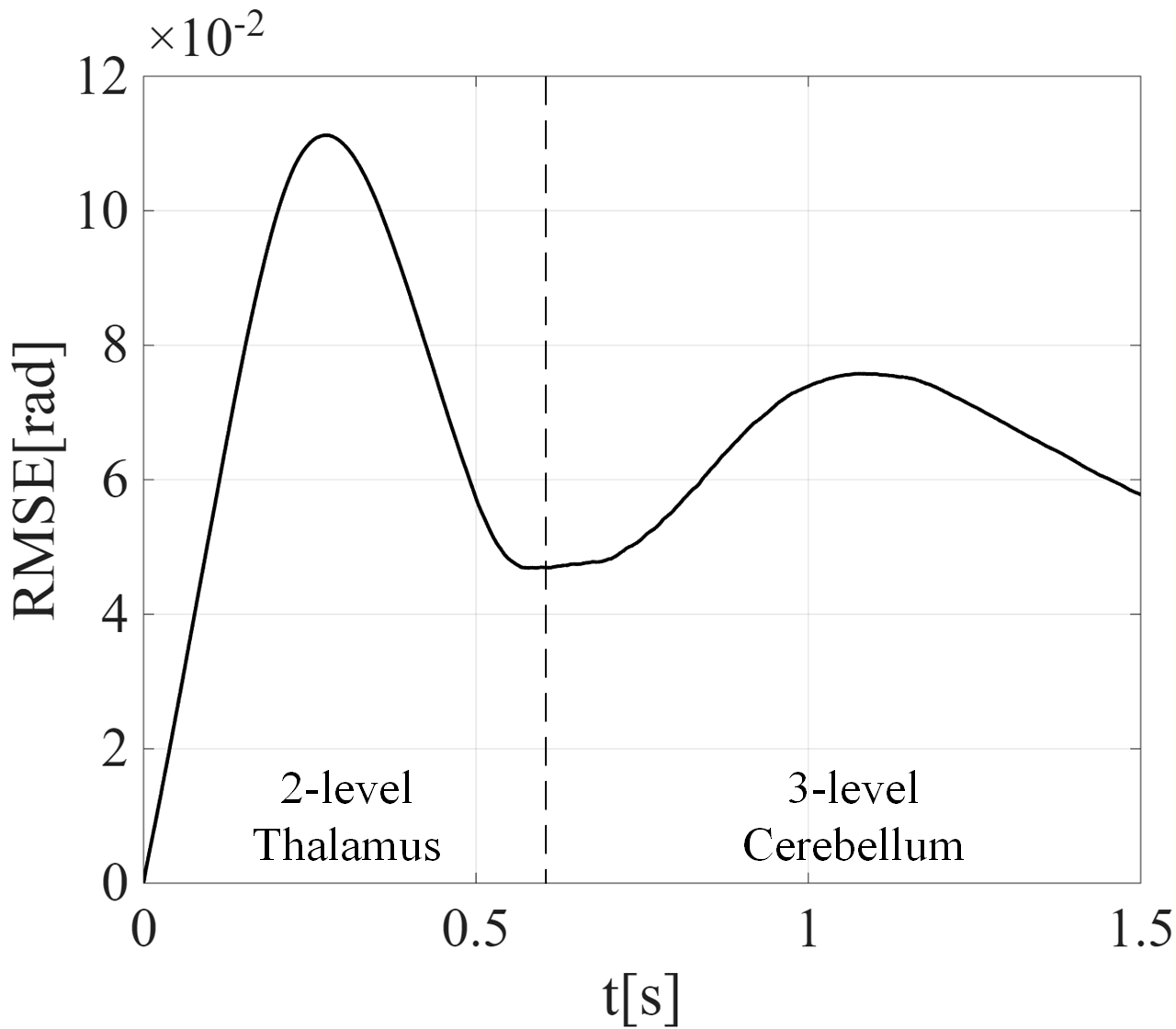}%
			\label{fig:vary_payload_error-e}}
		\subfloat[]{\includegraphics[width=0.25\linewidth]{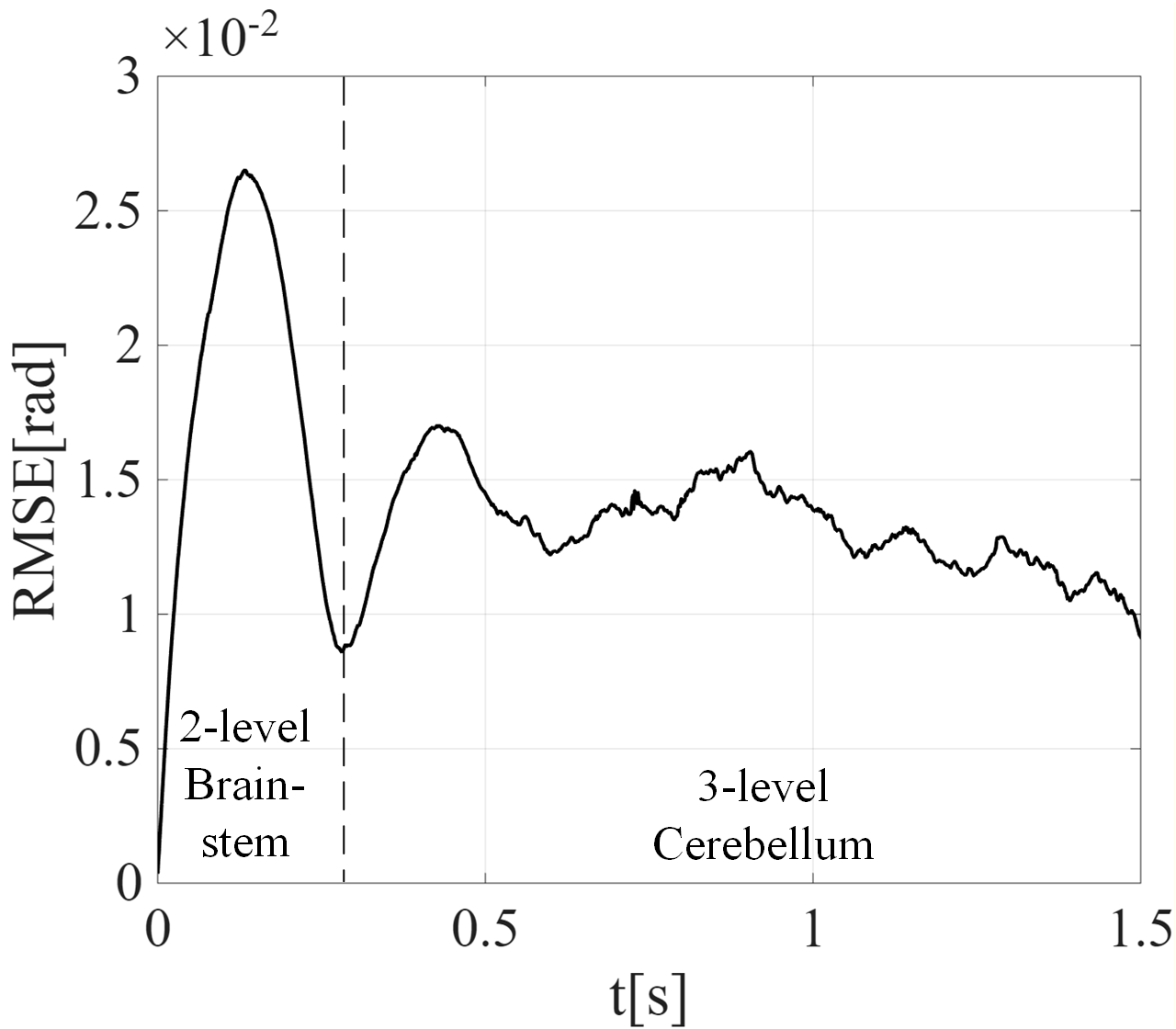}%
			\label{fig:vary_payload_error-f}}
		\subfloat[]{\includegraphics[width=0.25\linewidth]{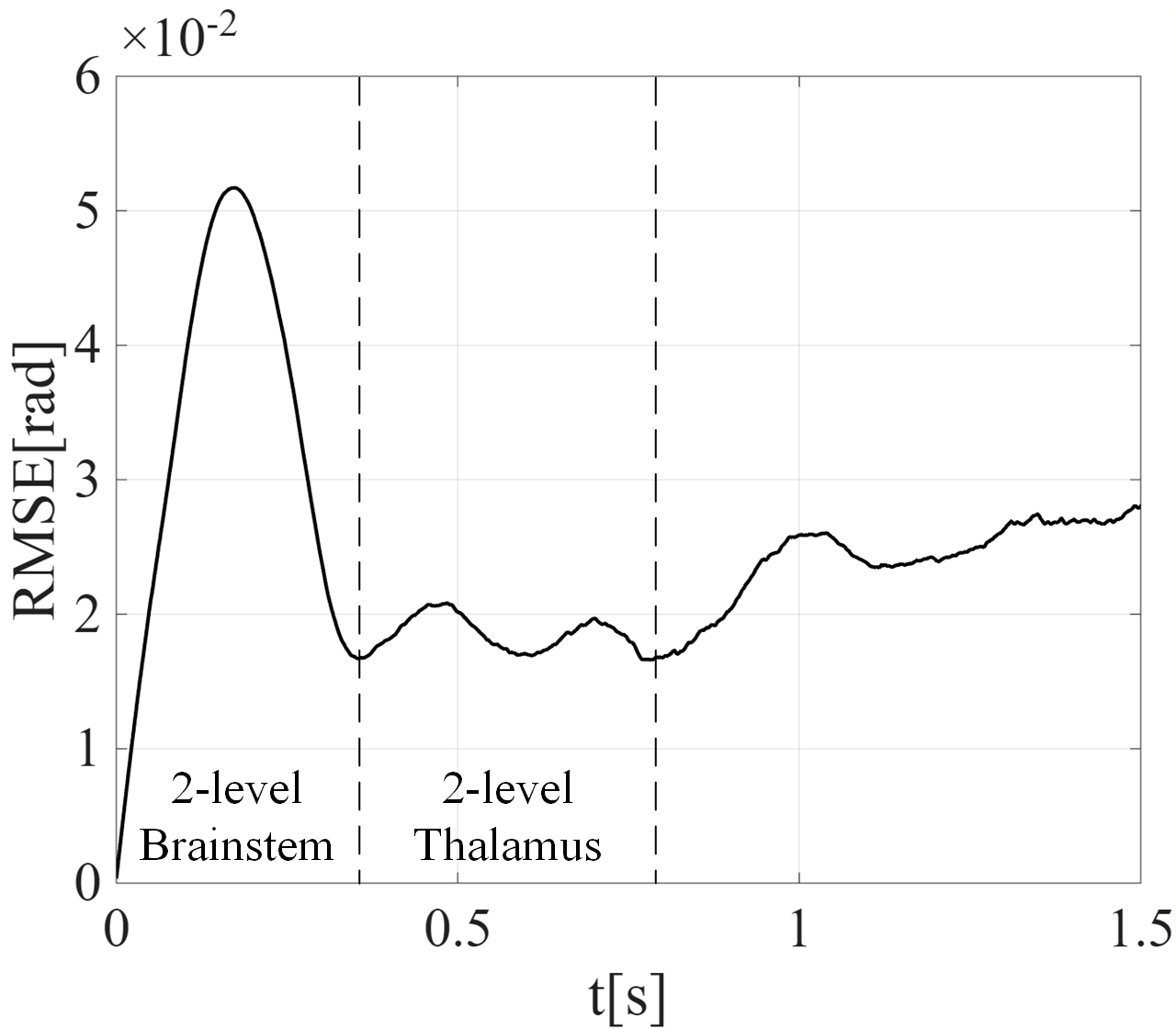}%
			\label{fig:vary_payload_error-g}}
	\end{minipage}
	\caption{Resilience performance under unstructured load changes. (a) Evolution of average RMSE across the three trajectories. (b) Expanded view of 0-1.5 s. (c) Expanded view of 12-14 s. (d) Expanded view of 24-25.5 s. (e) Expanded view of 0-1.5 s without the brainstem module. (f) Expanded view of 0-1.5 s without the thalamus module. (g) Expanded view of 0-1.5 s without the cerebellum module.}
	\label{fig:vary_payload_error}
\end{figure*}
Fig. \ref{fig:vary_payload_error} illustrates the resilience of the proposed framework under unstructured load variations. As shown in Fig. \ref{fig:vary_payload_error-a}, when a 1.5 kg load is applied at $t=0$ s, the framework recovers after a learning period of approximately 1.5 s. After 12 s of operation, an additional 2 kg load is introduced, and the system again recovers within about 2 s. When the added load is removed at 24 s, recovery is achieved within approximately 1.5 s. In contrast, cerebellum-inspired networks \cite{abadia2019robot, abadia2021cerebellar, doi:10.1126/scirobotics.adp2356} require nearly 100~s to adapt to a 500 g load change, while CBMC-V2 \cite{10769842} requires about 15 s to converge after initialization.

Furthermore, the resilience performance improves over time. Although the load change at 12 s is larger than that at 0 s, the resulting tracking error fluctuation during recovery is considerably smaller, as shown in Fig. \ref{fig:vary_payload_error-b} and Fig. \ref{fig:vary_payload_error-c}. Similarly, as illustrated in Fig. \ref{fig:vary_payload_error-d}, the fluctuation induced by load removal at 24 s is even smaller, despite the mass change being identical to that at 12 s. In contrast, the error fluctuations caused by load variations in \cite{abadia2019robot, abadia2021cerebellar, doi:10.1126/scirobotics.adp2356} do not diminish as operation time increases.

It can also be observed from Fig. \ref{fig:vary_payload_error-b}, Fig. \ref{fig:vary_payload_error-c}, and Fig. \ref{fig:vary_payload_error-d} that the recovery process following each load change consistently exhibits three distinct error peaks. These peaks indicate that the framework converges in a hierarchical manner. Specifically, the spinal cord module (first-level) operates at the real-time control frequency and corresponds to the response reflected at each data point of the curve; the brainstem and thalamus modules (second-level) account for the first two peaks; and the cerebellum module (third-level) is responsible for the final peak. This stepwise activation pattern is analogous to human neural responses to perturbations, namely M1 reflexes, M2 reflexes, and voluntary movements \cite{forgaard2015voluntary}.

To verify that these peaks are associated with the framework’s hierarchical structure, ablation experiments are conducted by selectively disabling individual modules. As shown in Fig. \ref{fig:vary_payload_error-e}, removing the learning capability of the brainstem module by fixing the feedback control coefficients reduces the number of peaks to two and delays the first peak. Similarly, as illustrated in Fig. \ref{fig:vary_payload_error-f}, disabling the thalamus module by fixing the motor pattern weighting coefficients eliminates the second peak, leaving only the responses associated with the brainstem and cerebellum modules. Finally, as shown in Fig. \ref{fig:vary_payload_error-g}, replacing the cerebellum network output with a fixed compensation torque suppresses the cerebellum module peak and weakens the thalamus module response, since it relies on cerebellum module outputs to optimize motor pattern integration.

In summary, these results demonstrate that the proposed framework exhibits strong and progressively improving resilience during operation. This capability arises from the synergistic interaction of its hierarchical structure and modular design. Removing any module, control level, or pathway leads to a noticeable degradation in performance.


\section{Practical results}\label{sec:experiment}
\begin{figure}[]
	\centering
	\includegraphics[width=2.5in]{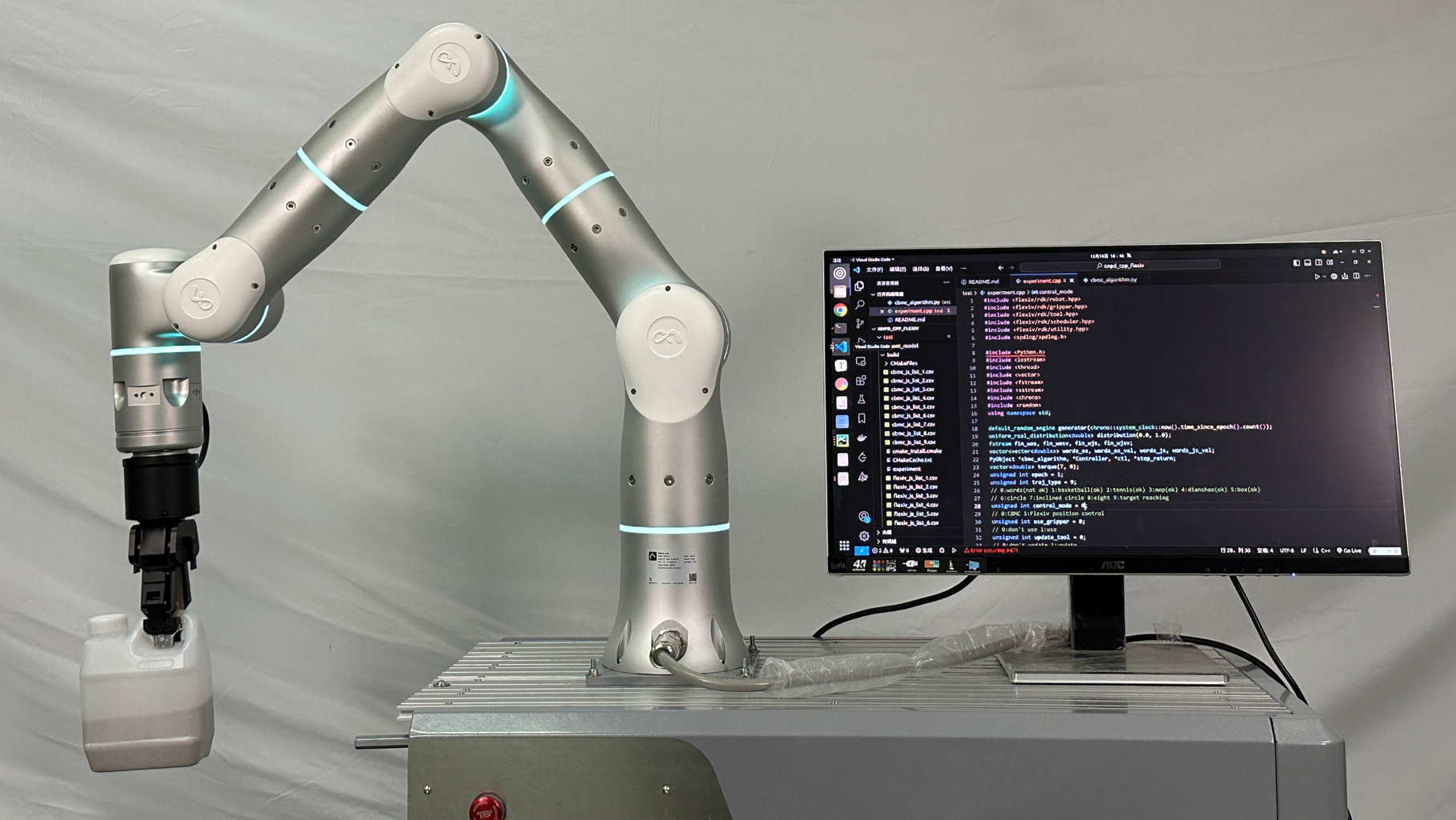}
	\caption{The experiment setting.}
	\label{fig:experiment}
\end{figure}

This section will present practical experiment results on a commercial robotic arm platform to demonstrate the precision and adaptation performance. The robotic arm platform is a Flexiv Rizon 4s equipped with a Flexiv GRAV gripper, as shown in Fig. \ref{fig:experiment}. The deep learning framework, SNN framework, control cycle of each level, and computing hardware are identical to that used in the simulation, and the Flexiv RDK version is 1.5.1.

To represent different tasks, five load masses, two load materials, and eight end-effector trajectories are designed. The load masses range from 1.5 kg (the gripper with an empty container) to 3.5 kg (the container with a 2 kg load), with increments of 500 g. The two load materials are sand and water. All trajectories are designed with a vertically downward orientation, with their definition and parameters provided in the Appendix. Among them, T1–T5 represent slower trajectories, whereas T6–T8 correspond to faster trajectories. The framework is evaluated on each trajectory ten times under different combinations of load mass and material.

In all experiments presented in this section, the trajectory tracking error is evaluated using the RMSE of the end-effector position and orientation. At each time step, the position and orientation RMSE are defined as
\begin{equation}
	\begin{aligned}
		&\text{RMSE}_p(t)=\\
		&\sqrt{(x_{d}(t)-x(t))^2+(y_{d}(t)-y(t))^2+(z_{d}(t)-z(t))^2}
	\end{aligned}
\end{equation}  
\begin{equation}
	\begin{aligned}
		&\text{RMSE}_o(t)=\\
		&\sqrt{(\theta_{d}(t)-\theta(t))^2+(\psi_{d}(t)-\psi(t))^2+(\phi_{d}(t)-\phi(t))^2}
	\end{aligned}
\end{equation} 
where $x_{d}, y_{d}, z_{d}, \theta_{d}, \psi_{d}, \phi_{d}$ denote the desired end-effector position in Cartesian space and the desired pitch, yaw, and roll angles, respectively, and $x, y, z, \theta, \psi, \phi$ denote the actual position in Cartesian space and the actual pitch, yaw, and roll angles, respectively. The average position and orientation RMSE of a trajectory are defined as 
\begin{align}
	\text{RMSE}_p=\frac{1}{L}\sum_{t=1}^{L}\text{RMSE}_p(t)\\
	\text{RMSE}_o=\frac{1}{L}\sum_{t=1}^{L}\text{RMSE}_o(t)
\end{align}
where $L$ represents the trajectory length.

\begin{figure*}[!t]
	\centering
	\includegraphics[width=0.8\linewidth]{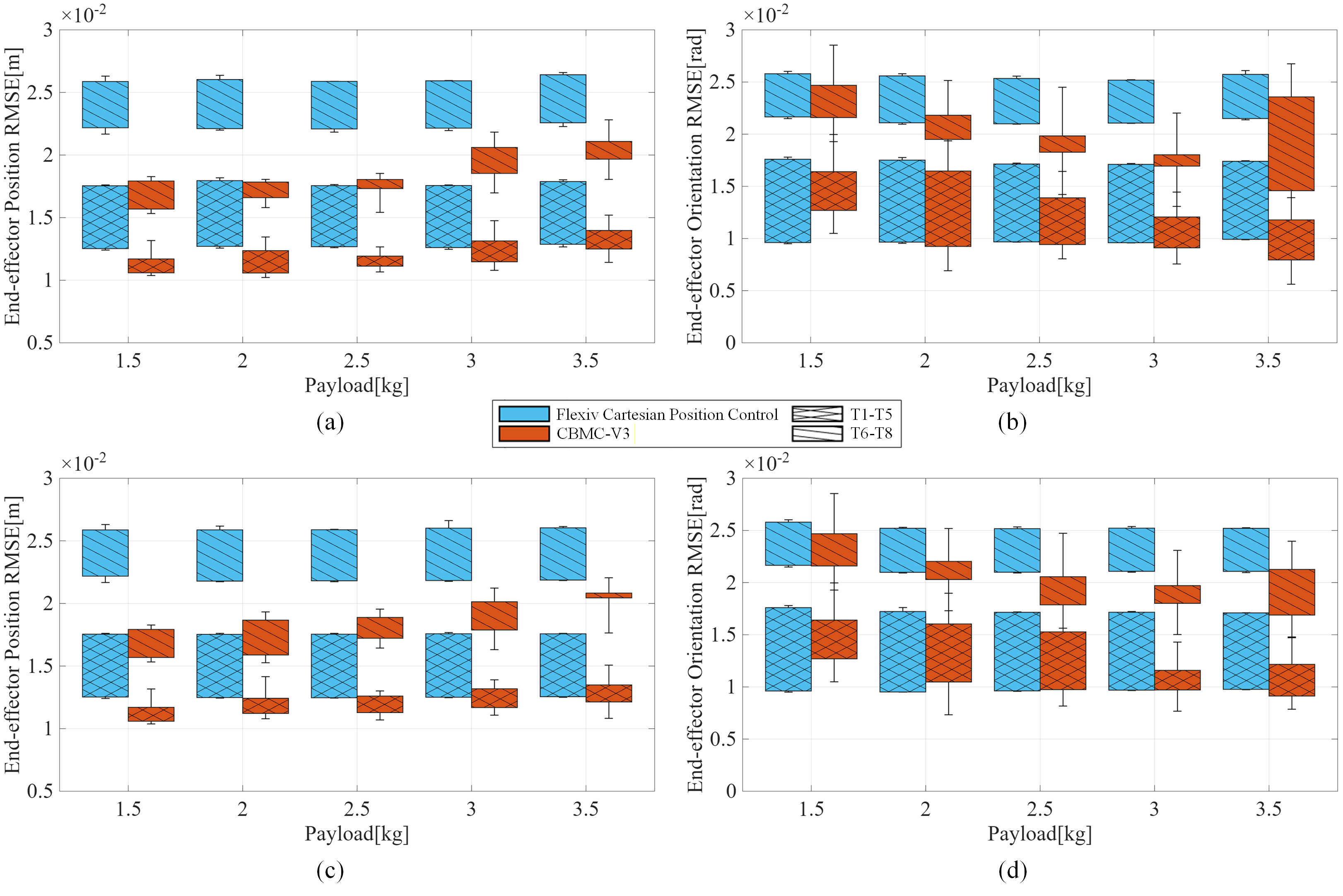}
	\caption{Precision and adaptation performance under different trajectory and load settings. (a) Position RMSE under sand load. (b) Orientation RMSE under sand load. (c) Position RMSE under water load. (d) Orientation RMSE under water load.}
	\label{fig:ex_result}
\end{figure*}

Fig. \ref{fig:ex_result} shows the trajectory tracking errors of the robotic arm under various trajectory and load conditions. Fig. \ref{fig:ex_result}a-b correspond to sand loads, while Fig. \ref{fig:ex_result}c-d correspond to water loads. The red and blue bars represent the proposed method and the factory-installed Flexiv Cartesian Position Control, respectively. Each bar indicates the range of average errors obtained by repeating each trajectory ten times under a given load condition, while the error bars denote the overall range of errors across all trajectories in the corresponding group. The slanted-hatched bars correspond to the fast trajectory group (T1–T5), whereas the cross-hatched bars represent the slow trajectory group (T6–T8).  Five observations can be drawn:

1. Precision: For all load masses, the proposed method consistently yields smaller average position and orientation errors, as indicated by the uniformly lower red bars compared with the blue bars. Under sand-load conditions, the Flexiv controller achieves average errors of $1.800\times10^{-2}$ m and $1.694\times10^{-2}$ rad across all trajectory and mass configurations, whereas the proposed method reduces these errors to $1.430\times10^{-2}$ m and $1.489\times10^{-2}$ rad, corresponding to reductions of 20.6\% and 12.1\%, respectively. Under water-load conditions, the Flexiv controller yields average errors of $1.779\times10^{-2}$ m and $1.681\times10^{-2}$ rad, while the proposed method achieves $1.442\times10^{-2}$ m and $1.519\times10^{-2}$ rad, resulting in reductions of 18.9\% and 9.6\%, respectively.

2. Adaptation to trajectory shape: Within each trajectory group, the proposed method exhibits greater consistency across different trajectory shapes, as the red bars under both hatching styles are consistently shorter than their blue counterparts. For sand-load conditions, the average variation ranges of the mean position and orientation errors for the Flexiv method are $0.441\times10^{-2}$ m and $0.596\times10^{-2}$ rad across all mass configurations, whereas those of the proposed method are reduced to $0.146\times10^{-2}$ m and $0.393\times10^{-2}$ rad, corresponding to reductions of 66.9\% and 34.1\%. For water-load conditions, the Flexiv method exhibits average variation ranges of $0.455\times10^{-2}$ m and $0.588\times10^{-2}$ rad, while the proposed method achieves $0.157\times10^{-2}$ m and $0.338\times10^{-2}$ rad, corresponding to reductions of 65.5\% and 39.4\%, respectively.

3. Adaptation to trajectory speed: The proposed method produces smaller tracking errors for both slow and fast trajectory groups. Notably, the improvement is more pronounced for the fast trajectories, as evidenced by the larger gap between the red and blue slanted-hatched bars compared with that between the cross-hatched bars. Under sand-load conditions, for the fast trajectory group, the proposed method reduces position and orientation errors by 29.4\% and 17.5\%, respectively, relative to the Flexiv method—exceeding the average improvements reported in Observation 1. Under water-load conditions, corresponding reductions of 28.5\% and 14.1\% are achieved. 


4. Adaptation to load mass: As the load mass increases, the position error of the proposed method increases, while the orientation error decreases; nevertheless, both remain consistently lower than those of the Flexiv method. This behavior is reflected by the upward trends of the red bars in Fig. \ref{fig:ex_result}a and Fig. \ref{fig:ex_result}c, and the downward trends in Fig. \ref{fig:ex_result}b and Fig. \ref{fig:ex_result}d. 


5. Adaptation to load material: Water introduces stronger dynamic variations due to its complex hydrodynamic behavior, thereby exerting a greater influence on control performance. Nevertheless, similar trends are observed when comparing Fig. \ref{fig:ex_result}c–d with Fig. \ref{fig:ex_result}a–b, and all of the above observations consistently apply to both load materials.


%

Taken together, the experimental results demonstrate that the proposed framework consistently outperforms the factory-installed method across a wide range of task conditions, including variations in trajectory shape, execution speed, load mass, and load material. The framework achieves higher precision and stronger adaptation to unstructured changes, with particularly advantages in high-speed scenarios.

\section{Conclusion and future work}\label{sec:conclusion}
In this paper, we propose a CNS-inspired, SNN-based motion control framework aimed at achieving agile manipulation in unstructured environments. The framework adopts a hierarchical architecture comprising five modules, three control levels, and two pathways, and is validated through both simulation studies and practical experiments. Taken together, the results demonstrate that the proposed method operates in a manner analogous to the human CNS and fulfills three out of the four key requirements of agile motion control outlined in Section \ref{sec:intro}, namely precision, adaptation, and resilience.

Energy efficiency is not explicitly analyzed in this paper, as it depends on multiple factors, including the deep learning and SNN frameworks, computational hardware, and the controlled robotic arm platform. In addition, orientation tracking performance is consistently inferior to position tracking performance, indicating room for further improvement. Moreover, adaptation to varying load masses can be further enhanced, as tracking precision degrades under changes in load mass. These aspects, along with a more thorough analysis of velocity tracking error, will be investigated in future work.


{\appendix[Practical experiment trajectory setting]
\begin{equation*}
	\text{T1:}\begin{cases}
		x=x_0+R_0\cos(\frac{2\pi t}{T_0}) \\
		y=y_0+R_0\sin(\frac{2\pi t}{T_0}) \\
		z=z_0
	\end{cases}
	\text{T2:}\begin{cases}
		x=x_1 \\
		y=y_1+R_1\sin(\frac{2\pi t}{T_1}) \\
		z=z_1+R_1\cos(\frac{2\pi t}{T_1})
	\end{cases}
\end{equation*}
\begin{equation*}
	\text{T3:}\begin{cases}
		x=x_2+R_2\cos(\frac{2\pi t}{T_2})\cos\theta_1 \\
		y=y_2+R_2\sin(\frac{2\pi t}{T_2}) \\
		z=z_2+R_2\cos(\frac{2\pi t}{T_2})\sin\theta_1
	\end{cases}
\end{equation*}
\begin{equation*}
		\text{T4:}\begin{cases}
		x=x_3+ 0.5 R_3\sin(\frac{4\pi t}{T_3})\\
		y=y_3+ R_3\cos(\frac{2\pi t}{T_3}) \\
		z=z_3+ z\sin(\frac{2\pi t}{T_3})
	\end{cases}
\end{equation*}
\begin{equation*}
	\text{T5:}\begin{cases}
		x=x_3+ 0.5 R_3\sin(\frac{4\pi t}{T_4})\\
		y=y_3+ R_3\cos(\frac{2\pi t}{T_4}) \\
		z=z_3+ z\sin(\frac{2\pi t}{T_4})
	\end{cases}
\end{equation*}
\begin{equation*}
	\text{T6:}\begin{cases}
		x=x_2+R_2\cos(\frac{2\pi t}{T_5})\cos\theta_1 \\
		y=y_2+R_2\sin(\frac{2\pi t}{T_5}) \\
		z=z_2+R_2\cos(\frac{2\pi t}{T_5})\sin\theta_1
	\end{cases}
\end{equation*}
\begin{equation*}
	\text{T7:}\begin{cases}
		x=x_4+R_4\cos(\frac{2\pi t}{T_6})\cos\theta_2 \\
		y=y_4+R_4\sin(\frac{2\pi t}{T_6}) \\
		z=z_4+R_4\cos(\frac{2\pi t}{T_6})\sin\theta_2
	\end{cases}
\end{equation*}
\begin{equation*}
	\text{T8:}\begin{cases}
		x=x_5+R_5\cos(\frac{2\pi t}{T_7})\cos\theta_2\cos\phi_1-R_4\sin(\frac{2\pi t}{T_7})\sin\phi_1\\
		y=y_5+R_5\cos(\frac{2\pi t}{T_7})\cos\theta_2\sin\phi_1+R_4\sin(\frac{2\pi t}{T_7})\cos\phi_1 \\
		z=z_5+R_5\cos(\frac{2\pi t}{T_7})\sin\theta_2
	\end{cases}.
\end{equation*}

\begin{table}
	\centering
	\caption{Trajectory parameters}
	\renewcommand\arraystretch{1.5}
	\begin{tabular}{cc|cc|cc}
		\toprule
		Parameter  & Value & Parameter & Value & Parameter & Value      \\
		\midrule
		
		$(x_0,y_0,z_0)$ & \makecell{(0.54,0.0,\\0.45) m} & $R_0$ & 0.14 m & $T_0$ & 4 s\\
		$(x_1,y_1,z_1)$ & \makecell{(0.6,0.35,\\0.0) m} & $R_1$ & 0.2 m & $T_1$ & 4 s\\
		$(x_2,y_2,z_2)$ & \makecell{(0.63,-0.11,\\0.3) m} & $R_2$ & 0.14 m & $T_2$ & 4 s\\
		$(x_3,y_3,z_3)$ & \makecell{(0.0,0.61,\\0.3) m} & $R_3$ & 0.14 m & $T_3$ & 4s\\
		$(x_4,y_4,z_4)$ & \makecell{(0.53,-0.11,\\0.4) m} & $R_4$ & 0.2 m & $T_4$ & 3s\\
		$(x_5,y_5,z_5)$ & \makecell{(0.53,-0.11,\\0.4) m} & $R_5$ & 0.2 m & $T_5$ & 2s\\
		$\theta_1$ & $-\frac{\pi}{6}$ rad & \multirow{2}{*}{$\phi_1$} & \multirow{2}{*}{$\frac{\pi}{3}$ rad} & $T_6$ & 3s \\
		$\theta_2$ & $\frac{\pi}{6}$ rad &  &   & $T_7$ & 3s\\
		
		\bottomrule
	\end{tabular}
	\label{tab:experiment traj}
\end{table}}

\bibliographystyle{IEEEtran}
\bibliography{refs}

@inproceedings{wang2023trajectory,
  title={Trajectory Tracking Control for Robot Manipulator Under Dynamic Environment},
  author={Wang, Yushi and Pang, Yanbo and Li, Qingkai and Cai, Wenhan and Zhao, Mingguo},
  booktitle={International Conference on Intelligent Robotics and Applications},
  pages={513--524},
  year={2023},
  organization={Springer}
}

@article{chaoui2009ann,
  title={ANN-based adaptive control of robotic manipulators with friction and joint elasticity},
  author={Chaoui, Hicham and Sicard, Pierre and Gueaieb, Wail},
  journal={IEEE Transactions on Industrial Electronics},
  volume={56},
  number={8},
  pages={3174--3187},
  year={2009},
  publisher={IEEE}
}

@article{he2017adaptive,
  title={Adaptive fuzzy neural network control for a constrained robot using impedance learning},
  author={He, Wei and Dong, Yiting},
  journal={IEEE transactions on neural networks and learning systems},
  volume={29},
  number={4},
  pages={1174--1186},
  year={2017},
  publisher={IEEE}
}

@article{wang2018adaptive,
  title={Adaptive neural network-based visual servoing control for manipulator with unknown output nonlinearities},
  author={Wang, Fujie and Liu, Zhi and Chen, CL Philip and Zhang, Yun},
  journal={Information Sciences},
  volume={451},
  pages={16--33},
  year={2018},
  publisher={Elsevier}
}

@article{liu2019adaptive,
  title={Adaptive neural network control with optimal number of hidden nodes for trajectory tracking of robot manipulators},
  author={Liu, Chengxiang and Zhao, Zhijia and Wen, Guilin},
  journal={Neurocomputing},
  volume={350},
  pages={136--145},
  year={2019},
  publisher={Elsevier}
}

@article{pham2020adaptive,
  title={Adaptive neural network based dynamic surface control for uncertain dual arm robots},
  author={Pham, Dung Tien and Nguyen, Thai Van and Le, Hai Xuan and Nguyen, Linh and Thai, Nguyen Huu and Phan, Tuan Anh and Pham, Hai Tuan and Duong, Anh Hoai and Bui, Lam Thanh},
  journal={International Journal of Dynamics and Control},
  volume={8},
  pages={824--834},
  year={2020},
  publisher={Springer}
}

@inproceedings{Abe2007,
author = {Abe, Yeuhi and da Silva, Marco and Popovi\'{c}, Jovan},
title = {Multiobjective Control with Frictional Contacts},
year = {2007},
isbn = {9781595936240},
publisher = {Eurographics Association},
address = {Goslar, DEU},
booktitle = {Proceedings of the 2007 ACM SIGGRAPH/Eurographics Symposium on Computer Animation},
pages = {249–258},
numpages = {10},
location = {San Diego, California},
series = {SCA '07}
}

@inproceedings{liu2011interactive,
  title={Interactive dynamics and balance of a virtual character during manipulation tasks},
  author={Liu, Mingxing and Micaelli, Alain and Evrard, Paul and Escande, Adrien and Andriot, Claude},
  booktitle={2011 IEEE International Conference on Robotics and Automation},
  pages={1676--1682},
  year={2011},
  organization={IEEE}
}

@INPROCEEDINGS{weighttans2011,  
author={Salini, Joseph and Padois, Vincent and Bidaud, Philippe}, 
booktitle={2011 IEEE International Conference on Robotics and Automation},   
title={Synthesis of complex humanoid whole-body behavior: A focus on sequencing and tasks transitions},   
year={2011},  volume={},  number={},  
pages={1283-1290},  
doi={10.1109/ICRA.2011.5980202}
}

@ARTICLE{Lee2012Intermediate,  
author={Lee, Jaemin and Mansard, Nicolas and Park, Jaeheung},  
journal={IEEE Transactions on Robotics},   
title={Intermediate Desired Value Approach for Task Transition of Robots in Kinematic Control},   
year={2012},  
volume={28},  number={6},  pages={1260-1277},  
doi={10.1109/TRO.2012.2210293}
}

@article{Escande2014,
author = {Adrien Escande and Nicolas Mansard and Pierre-Brice Wieber},
title ={Hierarchical quadratic programming: Fast online humanoid-robot motion generation},
journal = {The International Journal of Robotics Research},
volume = {33},
number = {7},
pages = {1006-1028},
year = {2014},
doi = {10.1177/0278364914521306}
}

@ARTICLE{Kim2019,  
author={Kim, Sanghyun and Jang, Keunwoo and Park, Suhan and Lee, Yisoo and Lee, Sang Yup and Park, Jaeheung},  journal={IEEE Robotics and Automation Letters},   title={Continuous Task Transition Approach for Robot Controller Based on Hierarchical Quadratic Programming},   year={2019},  volume={4},  number={2},  pages={1603-1610},  doi={10.1109/LRA.2019.2896769}}

@INPROCEEDINGS{han2021recursive,
  author={Han, Gang and Wang, Jiajun and Ju, Xiaozhu and Zhao, Mingguo},
  booktitle={2022 IEEE/RSJ International Conference on Intelligent Robots and Systems (IROS)}, 
  title={Recursive Hierarchical Projection for Whole-Body Control with Task Priority Transition}, 
  year={2022},
  volume={},
  number={},
  pages={11312-11319},
  doi={10.1109/IROS47612.2022.9981328}}

@article{carron2019data,
  title={Data-driven model predictive control for trajectory tracking with a robotic arm},
  author={Carron, Andrea and Arcari, Elena and Wermelinger, Martin and Hewing, Lukas and Hutter, Marco and Zeilinger, Melanie N},
  journal={IEEE Robotics and Automation Letters},
  volume={4},
  number={4},
  pages={3758--3765},
  year={2019},
  publisher={IEEE}
}

@article{pankert2020perceptive,
  title={Perceptive model predictive control for continuous mobile manipulation},
  author={Pankert, Johannes and Hutter, Marco},
  journal={IEEE Robotics and Automation Letters},
  volume={5},
  number={4},
  pages={6177--6184},
  year={2020},
  publisher={IEEE}
}

@article{salloom2020adaptive,
  title={Adaptive neural network control of underwater robotic manipulators tuned by a genetic algorithm},
  author={Salloom, Tony and Yu, Xinbo and He, Wei and Kaynak, Okyay},
  journal={Journal of Intelligent \& Robotic Systems},
  volume={97},
  pages={657--672},
  year={2020},
  publisher={Springer}
}

@article{hodgkin1952quantitative,
  title={A quantitative description of membrane current and its application to conduction and excitation in nerve},
  author={Hodgkin, Alan L and Huxley, Andrew F},
  journal={The Journal of physiology},
  volume={117},
  number={4},
  pages={500},
  year={1952},
  publisher={Wiley-Blackwell}
}

@article{burkitt2006review,
  title={A review of the integrate-and-fire neuron model: I. Homogeneous synaptic input},
  author={Burkitt, Anthony N},
  journal={Biological cybernetics},
  volume={95},
  pages={1--19},
  year={2006},
  publisher={Springer}
}

@article{stein1965theoretical,
  title={A theoretical analysis of neuronal variability},
  author={Stein, Richard B},
  journal={Biophysical journal},
  volume={5},
  number={2},
  pages={173--194},
  year={1965},
  publisher={Elsevier}
}

@article{bing2018survey,
  title={A survey of robotics control based on learning-inspired spiking neural networks},
  author={Bing, Zhenshan and Meschede, Claus and R{\"o}hrbein, Florian and Huang, Kai and Knoll, Alois C},
  journal={Frontiers in neurorobotics},
  volume={12},
  pages={35},
  year={2018},
  publisher={Frontiers Media SA}
}

@article{carrillo2008real,
  title={A real-time spiking cerebellum model for learning robot control},
  author={Carrillo, Richard R and Ros, Eduardo and Boucheny, Christian and Olivier, J-MD Coenen},
  journal={Biosystems},
  volume={94},
  number={1-2},
  pages={18--27},
  year={2008},
  publisher={Elsevier}
}

@article{abadia2019robot,
  title={On robot compliance: A cerebellar control approach},
  author={Abadia, Ignacio and Naveros, Francisco and Garrido, Jesus A and Ros, Eduardo and Luque, Niceto R},
  journal={IEEE transactions on cybernetics},
  volume={51},
  number={5},
  pages={2476--2489},
  year={2019},
  publisher={IEEE}
}

@article{abadia2021cerebellar,
  title={A cerebellar-based solution to the nondeterministic time delay problem in robotic control},
  author={Abad{\'\i}a, Ignacio and Naveros, Francisco and Ros, Eduardo and Carrillo, Richard R and Luque, Niceto R},
  journal={Science Robotics},
  volume={6},
  number={58},
  pages={eabf2756},
  year={2021},
  publisher={American Association for the Advancement of Science}
}

@article{georgopoulos1986neuronal,
  title={Neuronal population coding of movement direction},
  author={Georgopoulos, Apostolos P and Schwartz, Andrew B and Kettner, Ronald E},
  journal={Science},
  volume={233},
  number={4771},
  pages={1416--1419},
  year={1986},
  publisher={American Association for the Advancement of Science}
}

@book{gerstner2002spiking,
  title={Spiking neuron models: Single neurons, populations, plasticity},
  author={Gerstner, Wulfram and Kistler, Werner M},
  year={2002},
  publisher={Cambridge university press}
}

@article{LLOBERA2023190,
title = {Physics-based character animation and human motor control},
journal = {Physics of Life Reviews},
volume = {46},
pages = {190-219},
year = {2023},
issn = {1571-0645},
doi = {https://doi.org/10.1016/j.plrev.2023.06.012},
url = {https://www.sciencedirect.com/science/article/pii/S1571064523000775},
author = {Joan Llobera and Caecilia Charbonnier}
}

@article{annurev:/content/journals/10.1146/annurev-neuro-082321-025137,
   author = "Leiras, Roberto and Cregg, Jared M. and Kiehn, Ole",
   title = "Brainstem Circuits for Locomotion", 
   journal= "Annual Review of Neuroscience",
   year = "2022",
   volume = "45",
   number = "Volume 45, 2022",
   pages = "63-85",
   doi = "https://doi.org/10.1146/annurev-neuro-082321-025137",
   url = "https://www.annualreviews.org/content/journals/10.1146/annurev-neuro-082321-025137",
   publisher = "Annual Reviews",
   issn = "1545-4126",
   type = "Journal Article",
}

@article{merel2019hierarchical,
  title={Hierarchical motor control in mammals and machines},
  author={Merel, Josh and Botvinick, Matthew and Wayne, Greg},
  journal={Nature communications},
  volume={10},
  number={1},
  pages={5489},
  year={2019},
  publisher={Nature Publishing Group UK London}
}

@book{Halassa_2022,
   place={Cambridge}, 
   author={Michael M. Halassa},
   title={The Thalamus},  
   publisher={Cambridge University Press}, 
   year={2022}
}

@ARTICLE{8891809,
  author={Neftci, Emre O. and Mostafa, Hesham and Zenke, Friedemann},
  journal={IEEE Signal Processing Magazine}, 
  title={Surrogate Gradient Learning in Spiking Neural Networks: Bringing the Power of Gradient-Based Optimization to Spiking Neural Networks}, 
  year={2019},
  volume={36},
  number={6},
  pages={51-63},
  doi={10.1109/MSP.2019.2931595}
}

@INPROCEEDINGS{10769842,
  author={Pang, Yanbo and Li, Qingkai and Wang, Yushi and Zhao, Mingguo},
  booktitle={2024 IEEE-RAS 23rd International Conference on Humanoid Robots (Humanoids)}, 
  title={CBMC-V2: A CNS-inspired Framework for Real-time Robotic Arm Control}, 
  year={2024},
  volume={},
  number={},
  pages={121-128},
  doi={10.1109/Humanoids58906.2024.10769842}
}

@Article{robotics13090126,
AUTHOR = {Urrea, Claudio and Garcia-Garcia, Yainet and Kern, John},
TITLE = {Closed-Form Continuous-Time Neural Networks for Sliding Mode Control with Neural Gravity Compensation},
JOURNAL = {Robotics},
VOLUME = {13},
YEAR = {2024},
NUMBER = {9},
ARTICLE-NUMBER = {126},
URL = {https://www.mdpi.com/2218-6581/13/9/126},
ISSN = {2218-6581},
DOI = {10.3390/robotics13090126}
}

@article{henkes2024spiking,
  title={Spiking neural networks for nonlinear regression},
  author={Henkes, Alexander and Eshraghian, Jason K and Wessels, Henning},
  journal={Royal Society Open Science},
  volume={11},
  number={5},
  pages={231606},
  year={2024},
  publisher={The Royal Society}
}

@article{bennett1992time,
  title={Time-varying stiffness of human elbow joint during cyclic voluntary movement},
  author={Bennett, DJ and Hollerbach, JM and Xu, Y and Hunter, IW},
  journal={Experimental brain research},
  volume={88},
  pages={433--442},
  year={1992},
  publisher={Springer}
}

@article{paszke2019pytorch,
  title={Pytorch: An imperative style, high-performance deep learning library},
  author={Paszke, Adam and Gross, Sam and Massa, Francisco and Lerer, Adam and Bradbury, James and Chanan, Gregory and Killeen, Trevor and Lin, Zeming and Gimelshein, Natalia and Antiga, Luca and others},
  journal={Advances in neural information processing systems},
  volume={32},
  year={2019}
}

@article{
doi:10.1126/sciadv.adi1480,
author = {Wei Fang  and Yanqi Chen  and Jianhao Ding  and Zhaofei Yu  and Timothée Masquelier  and Ding Chen  and Liwei Huang  and Huihui Zhou  and Guoqi Li  and Yonghong Tian },
title = {SpikingJelly: An open-source machine learning infrastructure platform for spike-based intelligence},
journal = {Science Advances},
volume = {9},
number = {40},
pages = {eadi1480},
year = {2023},
doi = {10.1126/sciadv.adi1480},
URL = {https://www.science.org/doi/abs/10.1126/sciadv.adi1480},
eprint = {https://www.science.org/doi/pdf/10.1126/sciadv.adi1480}
}

@MISC{coumans2021,
author =   {Erwin Coumans and Yunfei Bai},
title =    {PyBullet, a Python module for physics simulation for games, robotics and machine learning},
howpublished = {\url{http://pybullet.org}},
year = {2016--2021}
}

@online{thalamus,
  title = {Thalamus},
  author = {},
  year = {2022},
  organization = {Cleveland Clinic},
  url = {https://my.clevelandclinic.org/health/body/22652-thalamus},
  urldate = {2022-3-30},
  }

@online{brainstem,
  title = {Brainstem},
  author = {},
  year = {2024},
  organization = {Cleveland Clinic},
  url = {https://my.clevelandclinic.org/health/body/21598-brainstem},
  urldate = {2024-6-12},
  }

@online{cerebellum,
  title = {Cerebellum},
  author = {},
  year = {2022},
  organization = {Cleveland Clinic},
  url = {https://my.clevelandclinic.org/health/body/23418-cerebellum},
  urldate = {2022-7-7},
  }

@article{sanger2000human,
  title={Human arm movements described by a low-dimensional superposition of principal components},
  author={Sanger, Terence David},
  journal={Journal of Neuroscience},
  volume={20},
  number={3},
  pages={1066--1072},
  year={2000},
  publisher={Society for Neuroscience}
}

@article{forgaard2015voluntary,
  title={Voluntary reaction time and long-latency reflex modulation},
  author={Forgaard, Christopher J and Franks, Ian M and Maslovat, Dana and Chin, Laurence and Chua, Romeo},
  journal={Journal of neurophysiology},
  volume={114},
  number={6},
  pages={3386--3399},
  year={2015},
  publisher={American Physiological Society Bethesda, MD}
}

@ARTICLE{8610238,
  author={Eckert, Peter and Ijspeert, Auke J.},
  journal={IEEE Transactions on Robotics}, 
  title={Benchmarking Agility For Multilegged Terrestrial Robots}, 
  year={2019},
  volume={35},
  number={2},
  pages={529-535},
  doi={10.1109/TRO.2018.2888977}}

@article{
doi:10.1126/scirobotics.adp2356,
author = {Ignacio Abadía  and Alice Bruel  and Grégoire Courtine  and Auke J. Ijspeert  and Eduardo Ros  and Niceto R. Luque },
title = {A neuromechanics solution for adjustable robot compliance and accuracy},
journal = {Science Robotics},
volume = {10},
number = {98},
pages = {eadp2356},
year = {2025},
doi = {10.1126/scirobotics.adp2356},
URL = {https://www.science.org/doi/abs/10.1126/scirobotics.adp2356},
eprint = {https://www.science.org/doi/pdf/10.1126/scirobotics.adp2356}}

@book{sinnatamby2013last,
  title={Last’s anatomy: regional and applied, 12th edn},
  author={Sinnatamby, Chummy S},
  volume={95},
  number={3},
  year={2013},
  publisher={Royal College of Surgeons}
}

@book{moore2018clinically,
  title={Clinically oriented anatomy},
  author={Moore, Keith L and Dalley, Arthur F},
  year={2018},
  publisher={Wolters kluwer india Pvt Ltd}
}

@article{sheppard2006agility,
  title={Agility literature review: Classifications, training and testing},
  author={Sheppard, Jeremy M and Young, Warren B},
  journal={Journal of sports sciences},
  volume={24},
  number={9},
  pages={919--932},
  year={2006},
  publisher={Taylor \& Francis}
}

\begin{IEEEbiographynophoto}{Yanbo Pang}
received the B.Eng. from Sun Yat-sen University in 2022 and the M.Sc. from Tsinghua University in 2025. He is currently an intern at Booster Robotics. His research interests include neuromorphic computing, neurorobotics and robotic arm control.
\end{IEEEbiographynophoto}
\vspace{-23pt}
\begin{IEEEbiographynophoto}{Qingkai Li}
received his B.S. from Tsinghua University in 2021. He is currently a Ph.D. candidate in the Department of Automation at Tsinghua University, with research interests spanning neuromorphic computing, bio-inspired algorithms, imitation learning, and whole-body control.
\end{IEEEbiographynophoto}
\vspace{-23pt}
\begin{IEEEbiographynophoto}{Mingguo Zhao} received the B.Eng., M.Sc. and Ph.D. from Harbin Institute of Technology in 1995, 1997 and 2001, respectively. He is currently a Full Professor with the Department of Automation and a director of the Beijing Innovation Center for Future Chips, Tsinghua University. His research interests include brain-inspired robot control and general robotic platform based on neuromorphic computing and hardware. 
\end{IEEEbiographynophoto}

\vfill

\end{document}